\documentclass[twoside,11pt]{article}

\usepackage{jmlr2e}
\usepackage{amssymb}
\usepackage[english]{babel}
\usepackage{amsfonts}
\usepackage{makeidx}
\usepackage{multirow}
\usepackage{caption}
\usepackage{amssymb,amsmath,amsfonts}
\usepackage[left,modulo]{lineno}
\usepackage{bm}
\usepackage{psfrag}
\usepackage{graphics}
\usepackage{fancyhdr}
\usepackage{eucal}
\usepackage[english]{babel}
\usepackage{ifpdf}
\usepackage{makeidx}
\usepackage{setspace}
\usepackage{float}
\usepackage[T1]{fontenc}
\usepackage{subfigure}
\usepackage{dsfont}
\usepackage[ruled, vlined, linesnumbered, longend]{algorithm2e}
\usepackage{graphicx, framed}
\usepackage{textcomp}

\usepackage{epstopdf}

\usepackage{tikz}
\usepackage{tikz-inet}
\usetikzlibrary{shapes,snakes,arrows,automata}

\newif\ifnotes\notestrue

\def\hgr#1{}

\newcommand{\wpuvi}{w^{+ \, (0)}_{u,v}}
\newcommand{\wnuvi}{w^{- \, (0)}_{u,v}}

\newcommand{\wuv}{w^{*}_{u,v}}
\newcommand{\wuvt}{w^{*\,(\tau)}_{u,v}}
\newcommand{\wuvtm}{w^{*\,(\tau-1)}_{u,v}}

\newcommand{\op}{w^+}
\newcommand{\om}{w^-}
\newcommand{\Rho}{\mathrm{P}}

\newcommand{\ben}{\begin{enumerate}}
\newcommand{\een}{\end{enumerate}}

\newcommand{\bc}{\begin{center}}
\newcommand{\ec}{\end{center}}

\newcommand{\bit}{\begin{itemize}}
\newcommand{\eit}{\end{itemize}}

\newcommand{\ds}{\displaystyle}
\newcommand{\beq}{\begin{equation}}
\newcommand{\eeq}{\end{equation}}

\newcommand{\vak}{\va^{(k)}}
\newcommand{\vbk}{\vb^{(k)}}

\newcommand{\ppij}{p^+_{i,j}}
\newcommand{\pnij}{p^-_{i,j}}

\newcommand{\lpi}{\lambda^+_{i}}
\newcommand{\lnni}{\lambda^-_{i}}

\newcommand{\load}{\varrho}
\newcommand{\loadi}{\varrho_i}

\newcommand{\Tpi}{T^+_i}
\newcommand{\Tni}{T^-_i}

\newcommand{\wpij}{w^+_{i,j}}
\newcommand{\wnij}{w^-_{i,j}}

\newcommand{\w}{\mathbf{w}}

\newcommand{\va}{\mathbf{a}}
\newcommand{\vb}{\mathbf{b}}

\newcommand{\x}{\mathbf{x}}

\newcommand{\vg}{\boldsymbol\gamma}

\newcommand{\State}{\mathbf{S}}

\newcommand{\Prob}{\mathds{P}}

\begin{document}

\title{A Tutorial about Random Neural Networks in Supervised Learning}

\author{
\name{Sebasti\'an Basterrech}\email Sebastian.Basterrech.Tiscordio@vsb.cz\\
\addr Department of Computer Science\\
V\v{S}B-Technical University of Ostrava\\
Ostrava-Poruba, Czech Republic\\
\AND
\name{Gerardo Rubino}\email Gerardo.Rubino@inriafr\\
       \addr INRIA - Rennes\\
Rennes, France
}

\editor{--}

\maketitle

\begin{abstract}Random Neural Networks (RNNs) are a class of Neural Networks (NNs)
that can also be seen as a specific type of queuing network.
They have been successfully used in several domains during the last 25 years,
as queuing networks to analyze the performance of resource sharing in many
engineering areas, as learning tools and in combinatorial optimization, where
they are seen as neural systems, and also as models of neurological aspects
of living beings.
In this article we focus on their learning capabilities, and more
specifically, we present a practical guide for using the RNN to solve
supervised learning problems.
We give a general description of these models using almost indistinctly the
terminology of Queuing Theory and the neural one.
We present the standard learning procedures used by RNNs, adapted from
similar well-established improvements in the standard NN field.
We describe in particular a set of learning algorithms covering techniques based on the
use of first order and, then, of second order derivatives.
%
We also discuss some issues related to these objects and present new
perspectives about their use in supervised learning problems.
The tutorial describes their most relevant applications, and also provides
a large bibliography.

\end{abstract}

\begin{keywords}
Neural Networks, Random Neural Networks, Supervised Learning, Pattern Recognition, G-networks
\end{keywords}

\newcommand{\trans}{\bm{\varrho}}
\newcommand{\emis}{\bm{\varepsilon}}
\newcommand{\colec}{\bm{\theta}}

\section{Introduction}
%
Supervised Learning is an area of the Machine Learning field that refers to a set of problems wherein the information is presented according to an outcome measurement associated with a set of input features.
The information is presented as a dataset of labeled samples.
The aim is \textit{``to learn''} the relationship between input and output features.
This learning process is done based on a set of examples in order to generate a learning model with the power of \textit{``generalising''}, this is to make \textit{``good''} predictions for new unseen inputs.
The research on \textit{Neural Networks (NNs)} is considered to have started with the work of Warren McCulloch and Walter Pitts in 1943~\citep{CullochPitts43}, and it has produced a rich literature with a strong concentration of papers in the $80$s and $90$s.
In the $80$s Rumelhart \textit{et al.} explored the relationship between \textit{Parallel Distributed Processing (PDP)} systems and various aspects of human cognition. The authors defined a general framework of a PDP system reactivating the research on connectionist models~\citep{RumelhartChapter2}.
The most popular PDP systems are NNs. In the last decades several books and journals have been dedicated to the research on NNs.
The interest in the NN area arises from both its theoretic aspects and its computational power for solving real problems.
NNs have been successfully applied in many different fields such as engineering, biology, pattern recognition, theoretical physics, applied mathematics, statistics, etc.

There are many types of NNs, and the related literature is huge. This article focuses on a particular class of NNs called \textit{Random Neural Networks} .
The RNN model was introduced by E. Gelenbe in 1989~\citep{Gel89:RNNPosNeg,Gelenbe1989a}.
RNNs are mathematical objects that combine features of both NNs and queueing models.
They been successfully employed in many types of applications: in learning problems, in optimization, in image processing, in associative memories, etc.
Here, we are specifically interested in the situations where the model is applied for solving supervised learning tasks.
A RNN is a PDP composed of a pool of interconnected nodes, which process and transmit information (signals) between them.
Each node is a simple processor and it is characterized by its state, a whole number.
The nodes receives two kinds of signals (negative and positive) from their neighbors or from outside.
When a negative signal arrives to a node, it produces an effect that can be related to neural inhibition, its state its decreased by one. The arrivals of positive signals provoke the opposite effect, the state is increased by one.
The fire of signals by the nodes is modeled by Poisson processes, and the pattern of connectivity among the neurons follows stochastic rules.

%
%

The design of the model was inspired from the biological behavior of neuron circuits in the neo-cortex. The model considers the following biological aspects: the action potentials in the form of spikes, the exchange of excitatory and inhibitory signals among the neurons, the synapses (weighted connections between two neurons), random delays between spikes, reduction of neuronal potential after firing, arbitrarily topology~\citep{Gel89:RNNPosNeg}.
%
%
%
The model has been also proven very powerful, from the computational viewpoint.
In~\citep{Gelenbe2004b} the authors shown that under certain algebraic hypothesis the RNN is an universal approximator.
Besides, it can be easily implemented in both software and hardware.
In order to apply the model for solving learning tasks, several learning algorithms have been adapted from the classic NN to RNNs, such as the Gradient Descent~\citep{Gel93:Learning} and Quasi-Newton methods~\citep{Baster09,Likas00}.
The number of applications of the model in the learning area is very large, but the model has been also applied to solve combinatorial optimization problems, such that the Traveling Salesman Problem or the Minimum Vertex Covering Problem~\citep{Gelenbe1992,Gelenbe1993d}.
%
%

\subsubsection*{Main contributions}
The first overview about RNN was presented in 2000~\citep{Kocak00}.
A survey about RNN focused on networking application and self-aware networks was introduced in~\citep{Sakellari:2010}.
Another general and helpful survey about RNN was presented in~\citep{Thimotheou10}, where the authors describe the main applications of RNNs, covering several topics including biology, reinforcement learning, and optimization problems.
In~\citep{Kocak11} the authors focused on RNN for solving learning problems, they identified some drawbacks of the RNN learning applications.
In addition, an extensive literature about RNN was presented in~\citep{Do11}.
%
%
In the 25th anniversary of the RNN model, we present this tutorial that contains the following contributions with respect to the previous published material.
\begin{itemize}
\item We introduce the model as a simple computational processor in a PDP framework, instead of using concepts coming from queueing systems.
%
%
Besides, we present a parallelism between this particular PDP and the model as belonging to the queueing area.
\item We provide a structured overview about the numerical optimization algorithms used for training RNNs.
We introduce algorithms that use the first derivative information of a quadratic cost function, such that the gradient descent type algorithms.
We then present Quasi-Newton methods that use the information of the second derivative of the cost function.
In this practical guide, all the algorithms used for training are shown in detail following a homogeneous format.
\item We present a critical review and new perspectives on RNN in supervised learning.
We discuss technical issues concerning stability problems in the model itself, as well as problems related to the parameters' optimization in the learning process.
We discuss some points related to the computational advantages of the model, as well as about its weaknesses and limitations.
The overview concludes with remarks concerning some new trends and future research lines.
%
%
%
\end{itemize}
In addition, this article presents an overview of some selected applications of the RNN in the supervised learning area.
In particular, we comment on two applications where the experimental results show a better performance of the model with respect to other techniques of the literature.

\subsubsection*{Organization of the article}
This article is structured as follows.
%
%
Section~2 formally describes the RNN model as a learning tool and in the framework of queueing theory.
%
Section 3 presents algorithms for training the RNN model.
It starts with a formal specification of the computational problems in supervised learning.
In Sec.~\ref{GeneralIntro} we give a general description of RNN in the learning context.
We present the Gradient Descent algorithm in Sec.~\ref{GDsection}, and we introduce second order optimization methods in~\ref{QN}.
We describe the following algorithms: the Broyden-Fletcher-Goldfarb-Shanno in Sec.~\ref{BFGSAsubsection}, the Davidon-Fletcher-Powell in Sec.~\ref{DFPAsubsection}, the Levenberg-Marquardt in Sec.~\ref{LMsubsection} and one variation of it in Sec.~\ref{SectionLM-AM}.
We present a critical review about the RNN model for solving learning problems in Sec.~\ref{CriticalReview}.
%
Section~\ref{OverviewRNNsAplications} presents an overview of applications.
We conclude and present new research trends in Sec.~\ref{conclusionSection}.

\newcommand{\mmone}{\textit{M/M/1 }}

\section{The Random Neural Network model}
\label{RNNmodel}
This Section formally introduces the RNN model. It has four parts. First, we describe a single neuron (Random Neuron) as an elementary processor. Second, we present the RNN as a system composed by interconnected neurons.
Third, we review the model in the framework of queuing networks.
The section ends introducing the different topologies and structural concepts of the RNN.

\subsection{Random Neuron (RN)}
\begin{figure}[htbp]\centering
\fbox{\includegraphics[width=0.5\textwidth]{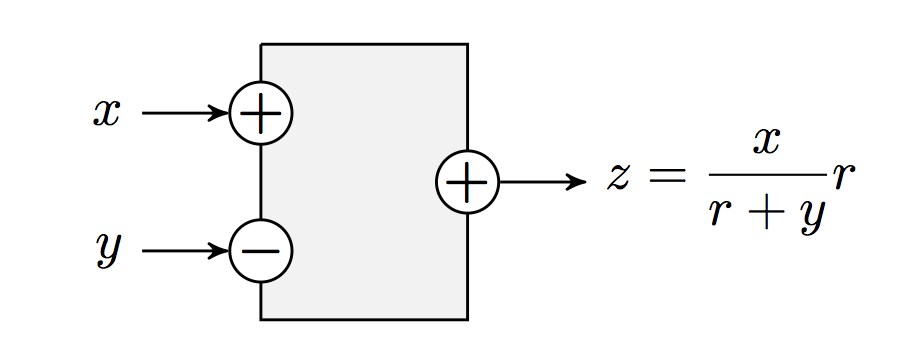}}
%
\caption{A zoom on a random neuron (RN) seen as a ``black-box'' system; the
inputs are the reals $x, y \geq 0$; the parameter is the rate $r > 0$,
and the output is the real $z$; we say that the first input variable $x$
is connected to the \textit{positive} input port of the RN (depicted `$+$') and
the second input variable $y$ to the \textit{negative} input port (depicted `$-$');
the output port is also said to be positive (and it is depicted `$+$' in the figure)
}
\label{f:rn}
\end{figure}

A \textit{Random Neuron (RN)} is a real parametric function of two real variables,
with a real parameter called the neuron's \textit{rate}. The input variables are
assumed to be non-negative. The rate is positive. If $x \geq 0$ is the first input
variable, $y \geq 0$ is the second one, and if $r > 0$ is the rate of the neuron, then the output is the real $z$ given by the expression
\begin{equation}
\label{z}
    z = \frac{ x }{ r + y } r.
\end{equation}
See that a RN is characterized by its rate $r$.
We can see the neuron as an input-output system with two ``input ports'',
one for $x$ and the other one for $y$, and one output port for $z$.
The ports associated with the output and with the first input value are
called \textit{positive}; the input port corresponding to
the second input variable $y$ is called \textit{negative}
(the reason for this is explained later), but all the variables involved
are non-negative real numbers.
Figure~\ref{f:rn} shows a neuron as an input-output device.
When $x \geq r + y$ we say that the neuron is \textit{saturated}.

The output value $z$ is seen as a measure of the activity of the neuron
(as in most input-output systems). As such, see that $z$ is increasing
in~$x$ and decreasing in~$y$. In real neurons, which also are input-output systems,
the input signals belong to two types, excitatory signals, which are those
contributing to the neuron's activity measured by its output (the higher the
excitatory signal, the higher the neuron's activity) and inhibiting inputs
playing the opposite role. This is why we call positive the signals arriving
at the `$+$' input port, and negative those arriving at the `$-$' one.

We will say that a RN is \textit{controlled} if its output $z$ is
modified according to the rule
\begin{equation}
\label{zCtrld}
    z = \min\left\{ \frac{ x }{ r + y }, \, 1 \right\}r.
\end{equation}
So, in this case the RN's output is always less than or equal to its rate,
and it is equal to its rate when the neuron is saturated.
In the case of the initial definition~(\ref{z}), the neuron is said to be \textit{uncontrolled}.

\subsection{Random Neural Network (RNN)}

A \textit{Random Neural Network} (RNN) is a network composed of $N$ interconnected RNs,
that implements a function from $\mathbb{R}^{2N}$ into $\mathbb{R}^O$,
for some $1 \leq O \leq N$, in the following way.
We are given $N$ RNs denoted $1, 2, \ldots, N$ (that is, we are given $N$
strictly positive reals $r_1, r_2, \ldots, r_N$), and two $N \times N$
matrices denoted by
$\mathbf{P}^+ = (p^+_{ij})$ and $\mathbf{P}^- = (p^-_{ij})$,
whose components are probabilities.
Both matrices \textit{and their sum} are substochastic, that is,
for any of their rows, say the~$i$th one, we have
    $$\sum_{j=1}^N \Big( p^+_{ij} + p^-_{ij} \Big) \leq 1.
    $$
Also, for at least one of the neurons,
we have $\sum_{j=1}^N (p^+_{ij} + p^-_{ij} ) < 1$.
The neurons~$i$ for which $\sum_{j=1}^N (p^+_{ij} + p^-_{ij} ) < 1$
are called \textit{output} neurons. We denote by $O$ their number
(so, $1 \leq O \leq N$).
The network outside is often referred to as the neuron's \textit{environment} with which the system operates~\citep{RumelhartChapter2}.

Let us denote the~$2N$ input variables of the network as
$x_1,\ldots,x_N,y_1,\ldots,y_N$. Then, the output of the network
is the set of outputs of each of its output neurons.
We need only to specify how are determined
the inputs to the $N$ RNs (the outputs are given by the previously described
rules, in the uncontrolled or controlled cases).
Let us call $u_i$ (respectively $v_i)$ the
positive (respectively negative) input to neuron~$i$. Then, the following
equations must be satisfied:
    $$u_i = x_i + \sum_{j=1}^N z_j p_{ji}^+,
    \qquad
        v_i = y_i + \sum_{j=1}^N z_j p_{ji}^-.
    $$
In words, the fraction $p_{ji}^+$ of the output $z_j$ of neuron~$j$
adds to the positive input $u_i$ to neuron~$i$, and the
fraction $p_{ji}^-$ of the output $z_j$ of neuron~$j$
adds to the negative input $v_i$ to~$i$.
Of course, this means that the reals $z_1,\ldots,z_N$ must satisfy the
non-linear system of equations
    $$z_i = \frac{\ds x_i + \sum_{j=1}^N z_j p_{ji}^+}%
        {\ds r_i + y_i + \sum_{j=1}^N z_j p_{ji}^-} r_i,
        \qquad i = 1, 2, \ldots, N.
    $$
This needs some technical discussions about the existence and unicity
of solutions to this system, as we will see below.

Observe that if we define
\begin{equation}
\label{probOutput}
    d_{i} = 1 - \sum_{j=1}^N \Big( p^+_{ij} + p^-_{ij} \Big),
\end{equation}
we have $0 \leq d_i \leq 1$, and that
neuron $i$ is an output neuron when $d_i > 0$. We can also say that
the network of neurons sends the part $d_i z_i$ of $z_i$ through the output
port of~$i$.

\textbf{Observation: } in general in the learning applications, we use a RNN with $N$ neurons as a function from $\mathbb{R}^I$ to $\mathbb{R}^O$ where $I < 2N$ or even $I < N$, by setting $2N-I$ of the standard $2N$ input variables to a fixed value (typically to~$0$).
We will see soon this frequent situation.
An important particular case covering all the applications done so far
for these objects as learning tools is as follows. The network with $N$
neurons implements a function with $I \leq N$ input variables and $O\leq N$ output variables.
The input variables are denoted by $x_1,\ldots,x_I$, which are all connected to the positive port of $I$ neurons called \textit{input} neurons.
In other words, no input variable is connected to a negative port.
The function output is the set of outputs generated by the $O$ output neurons.
A group of neurons can have no interactions with the environment (when $I+O<N$).
We call those units \textit{hidden} neurons. Note that a neuron can be both an input and an output one.

\subsection{A queueing view of the Random Neural Networks}
The RNN method has been used with two different interpretations both referring to exactly the same mathematical model. One is the already described type of interconnected RNs.
Another one is a type of queueing systems called G-queues and G-networks.
The first interpretation is often employed in the Machine Learning contexts and the second one is applied in Performance Evaluation, for example.

We begin by describing a single queue where customers arrive according to a Poisson process, say with rate $\lambda > 0$, and \textit{service times} are exponentially distributed with parameter $r > 0$.
It is assumed that service times are mutually independent and that they
are also independent of the inter-arrival times.
This server queue is named \textit{M/M/1} queueing model~\citep{Kendall53}.
At any time~$t$ the state of the system $S(t)$ is the number of customers present in the queue.
The queue storage capacity is infinite.
\newcommand{\limite}{\ds{\lim_{t \rightarrow \infty}}}
The stochastic process $\{S(t), \; t \geq 0\}$ is a continuous time homogeneous Markov process on the non-negative integers.
We define the \textit{utilization factor} of the queue as the ratio $\load= \lambda/r$.
When the process is ergodic ($\load< 1$), the steady-state is given by
\begin{equation}
    p(k) = \limite \Prob( S(t) = k ) = \load^k (1-\load).
\end{equation}

A Jackson queueing network consists of $N$ interconnected queues with the following characteristics. For each queue $i$ the service time is exponentially distributed with rate $r_i$.
When a customer completes the service at queue~$i$, it will either move to queue~$j$ with routing probability~$p_{ij}$ or leave the network with probability $d_i$
($d_i = 1-\sum_{j=1}^{N} p_{ij}$).
Customers arrive from the environment to queue~$i$ according to a Poisson process with rate $\lpi$.
At any time $t$, the system state is the vector $\State(t)=\left(S_1(t),\ldots,S_N(t) \right)$, where $S_i(t)$ denotes the number of customers in queue $i$ at time $t$.
The assumptions about the independence among the processes can be summarized as follows:
\bit
\item arrival processes, service processes and switching (routing)
processes are independent of each other;
\item at each server, the service times are independent of each other;
\item at each switching point, the successive switching results are independent of each other.
\eit

We define $T_i$ as the mean throughput at queue $i$.
In order to avoid a trivial case, we assume that at least one of the $\lpi$'s is non-zero (strictly positive).
In addition, assuming that the system is irreducible (for any two nodes $i$ and $j$ in the Markovian graph there exists a path from $i$ to $j$), and in equilibrium, $T_i$ for all $i$ can be determined by solving the \textit{flow balance equations}:
%
\begin{equation}
\label{flow}
T_i=\lpi+\ds{\sum_{j=1}^{N}T_jp_{ji}}.
\end{equation}
The strongly connected property of the Markovian graph implies that exists an  unique (and strictly positive) solution.
%
%
The utilization factor of queue $i$ is given by $\loadi= T_i/r_i$.
%
%

A G-network (or equivalently, an RNN) is an extension of a Jackson's network where there is a new entity in the system: \textit{negative customers}.
As in the previous network, in a G-network there are Poisson arrivals, probabilistic routing among the queues, exponential service rates and usual independence among the corresponding stochastic processes.
%
There are two types of customers in the system, positive ones that operate as we defined for the Jackson network, and the negative ones that operate as follows.
When a negative customer arrives at a non-empty queue, it destroys a positive customer in this queue, if any, and disappears.
If there are no customers in the queue, a negative customer does not operate, it just disappears from the system.
In several works negative customers are referenced as \textit{signals}, thus there are two entities, customers (positive customers) and signals (negative customers).

In~\citep{Gel89:RNNPosNeg,Gel91:ProdFormQNet} Gelenbe  shows that, in an equilibrium situation, the~$\loadi$s  satisfy the following flow balance equations:
    \begin{equation}\label{rhos}
    \mbox{for each node~$i$, } 
        \qquad \loadi = \frac{\Tpi}{r_i + \Tni},
    \end{equation}
    \begin{equation}\label{T+}
    \mbox{for each node~$i$, }
        \qquad \Tpi = \lpi + \sum_{j=1}^N \load_j r_jp^+_{ji},
    \end{equation} and
    \begin{equation}\label{T-}
    \mbox{for each node~$i$, }
        \qquad \Tni = \lnni + \sum_{j=1}^N \load_j r_jp^-_{ji},
    \end{equation}
with the supplementary condition that, for all neuron~$i$,
we have $\loadi < 1$.
An important result associated with open Jackson networks and with G-networks is called the \textit{product form theorem}.
Gelenbe proved that under Markovian assumptions G-networks have a product form equilibrium distribution. This means that the joint equilibrium distribution of the queue states is the product of the marginal distributions. For more details see~\citep{Gel89:RNNPosNeg}.

\begin{figure}[h]
\begin{center}
\fbox{
\includegraphics[width=0.9\textwidth]{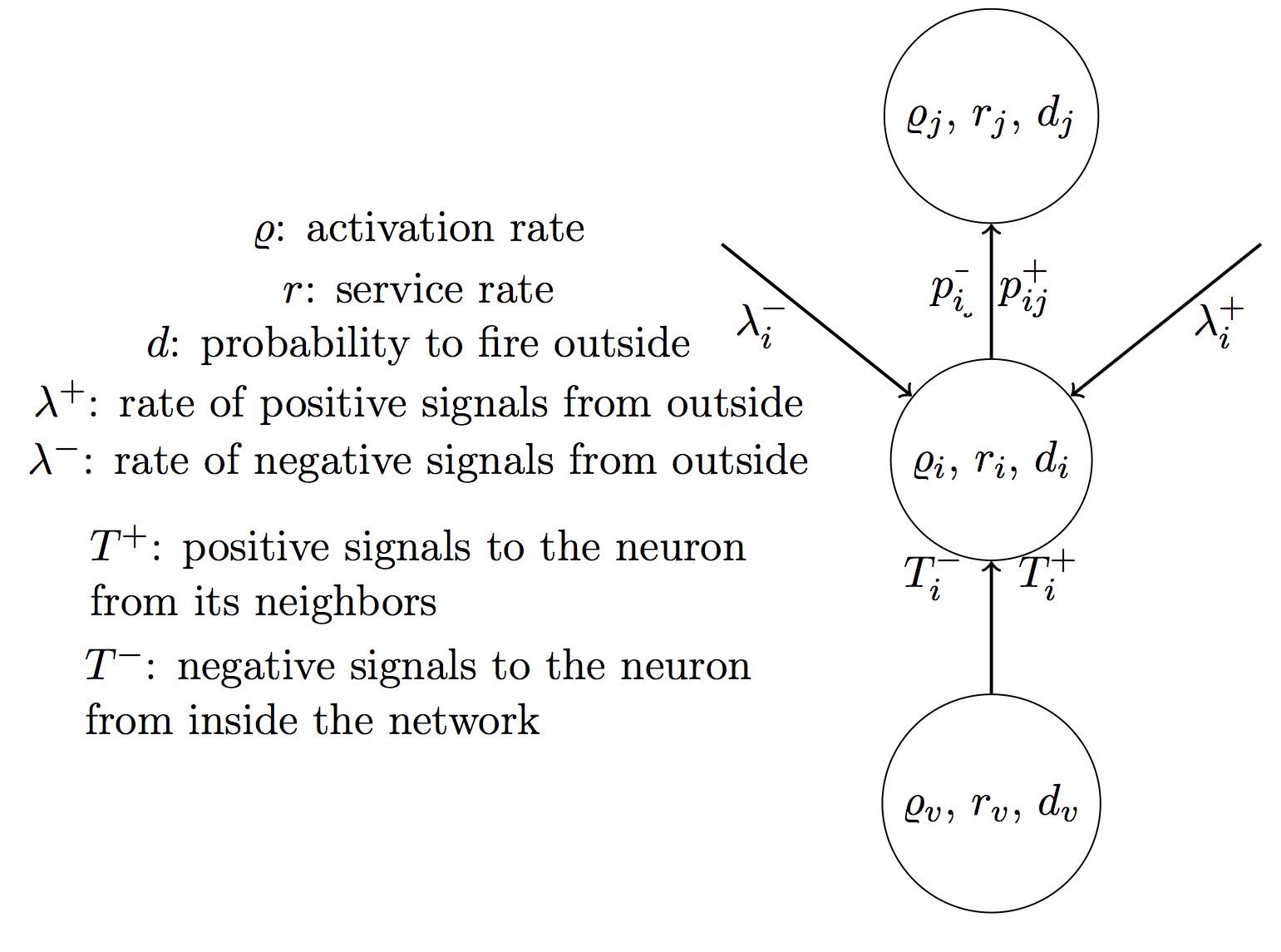}
}
\caption{\label{RNparameters} A representation of a RN. The figure shows the main parameters involved in a RN embedded in a network.
}
\end{center}
\end{figure}

\textbf{Observation:}  Let us unify the notation that will be used through this article.
So far we introduced the RNN as a function, next we presented the concept using a queueing point of view.
In the rest of the article, we follow the most often used notation presented in~\citep{Gel89:RNNPosNeg}.
Let $N$ be the number of interconnected neurons. For each neuron $i$ its service rate is denoted by $r_i$, the value at its positive port is denoted by $T_i^+$ and to the negative port is $T_i^-$. The positive input value $\lambda_i^+$ (the Poisson rate of the customers coming from outside), the negative input value $\lambda_i^-$ (the Poisson rate of the negative customers coming from outside), and the probability to send information to the environment denoted by $d_i$ characterize the interaction of $i$ with outside. The output of neuron $i$ is its activation rate $\load_i$.
The connections between two neurons $i$ and $j$ are given by the probabilities $\ppij$ and $\pnij$.
Figure~\ref{RNparameters} shows the main parameters involved in a RNN.
We will introduce in our notation the concept of weights. For any two neurons~$i$ and $j$, they are defined as: $\wpij=r_i\ppij$ and $\wnij=r_i\pnij$. The first one is called \textit{positive weight} and the second one is called \textit{negative weight}.
Note that the weights are, by definition, positive reals.
In the context of NNs, the traditional notation used for the weight connection (direct edge) between the nodes $i$ to $j$ is often denoted as $(j,i)$.
%
In the RNN context, the reverse order is traditionally used. This originates in the first paper about supervised learning with RNNs~\citep{Gel93:Learning}.
%

%

%
%
%
%
\newcommand{\sumi}{\displaystyle{\sum_{i\in\mathcal{I}}}}
\newcommand{\sumh}{\displaystyle{\sum_{h\in\mathcal{H}}}}
\newcommand{\sumo}{\displaystyle{\sum_{o\in\mathcal{O}}}}
\subsection{The network topology}
\label{ArchitectureNet}

So far, we defined the RNN as a parallel distributed system composed of simple processors (RNs).
Therefore, the network is a graph where the RNs are their nodes; the existence of an arc between two nodes is given by certain probability.
The two most common topologies of networks are \textit{multi-layer feedforward} and \textit{recurrent} networks.

\subsection{Feedforward topology}
\begin{figure}[h]
\begin{center}
\fbox{
\includegraphics[width=0.9\textwidth]{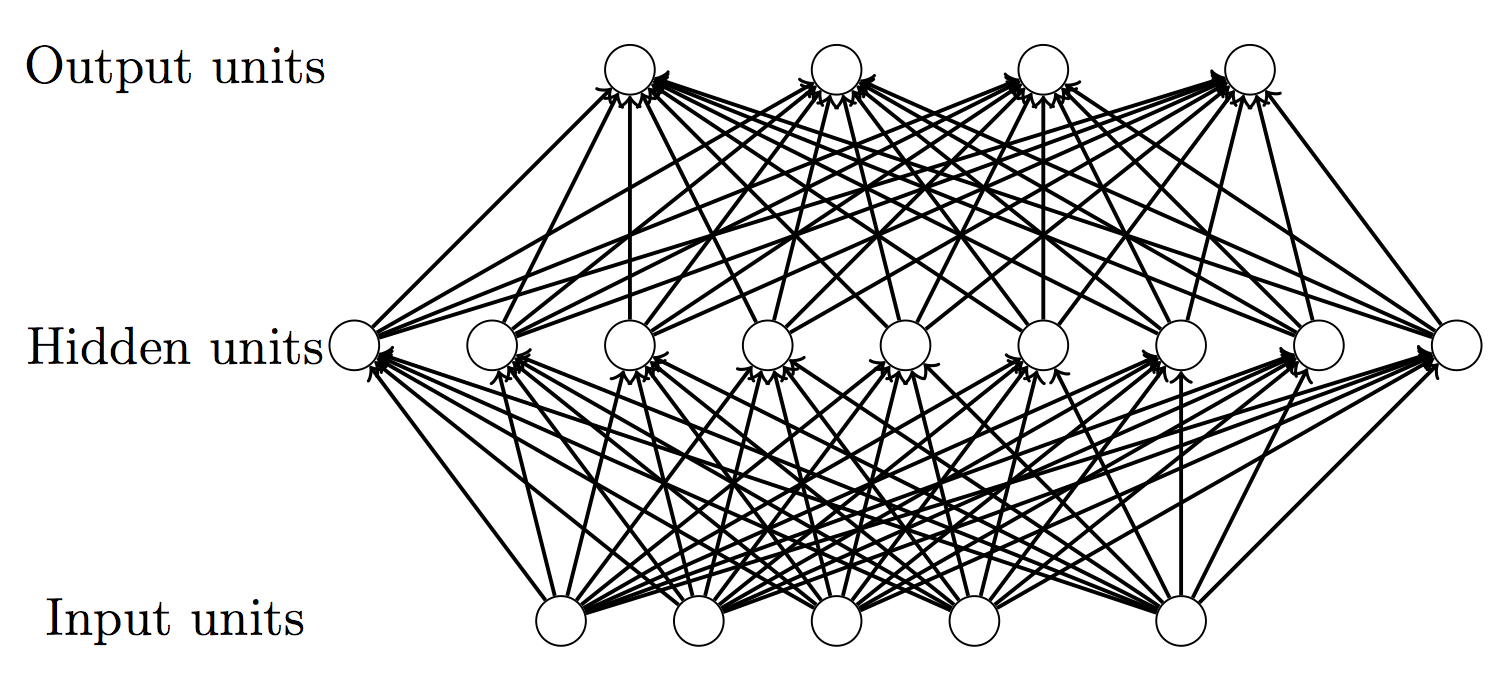}
}
\caption{\label{feedforwardDiagram} A representation of a Feedforward Neural Network.  The figure shows a network with a single hidden layer. The flow of information is from the the input neurons through the output ones.
In this example there are 5 input neurons full connected to $9$ hidden neurons, and the hidden neurons are full connected with $4$ output neurons. A network with this topology is used for mapping a relationship from a $5$-dimensional space into a $4$-dimensional space.
}
\end{center}
\end{figure}

We start describing the feedforward case.
The identifying property is that there are no cyclic connections among the neurons,
no circuits in the (directed) graph.
The architecture of the graphs consists of multiple layers of neurons in a directed graph.
There are three types of layers popularly known as \textit{input}, \textit{hidden} and  \textit{output} layers.
The neurons can have only connections in a forward direction, from the input neurons to the output neurons, traveling through the hidden ones.
Only neurons belonging to the input and to the output layers can exchange information with the environment.
The activity rate for each output neuron is computed using a forward propagation procedure.
A representation of a feedforward network with one hidden layer is illustrated in Figure~\ref{feedforwardDiagram}.

The feedforward case has been widely used in supervised learning due to the fact that training process is much faster than in the recurrent case.
Besides, the feedforward networks are easier to analyze than networks with recurrent topologies.
One advantage is that the non-linear system of equations~(\ref{rhos}), (\ref{T+}) and (\ref{T-}) can be formally solved.
Then, we can express the activity rate of the output units as functions of the inputs variables of the system.
Let $I$ be the number of input neurons, $H$ is the number of hidden neurons and let $O$ be the number of output neurons.
We arbitrary index the input neurons from $1$ to $I$,  the hidden neurons from $I+1$ to $I+H$ and the output neurons from $I+H+1$ to $I+H+O=N$.
We can compute the activity rate of the neurons using a forward procedure as follows.
At the first step, we compute the activity rate of the input neurons, next the activities  of the hidden neurons and finally those of the output neurons.
%
%
Input neurons are the only ones that receive signals from the environment; so we set $\lpi = \lnni = 0$ for all $i\in[I+1,N]$.
The activity rates are given by the following explicit expressions:
$$
\load_i = \frac{ \lpi }{ r_i + \lnni },\quad \forall i \in {[1,I]},
\qquad
\load_h = \frac{\ds{ \sum_{i=1}^{I} \load_i w^+_{i,h} }}%
{\ds{r_h + \sum_{i=1}^{I} \load_i w^-_{i,h}}},\quad \forall h \in [I+H+1,N],
$$
and
$$
\label{loadOutputFF}\qquad
        \load_o = \frac{\ds{\sum_{i=I+1}^{I+H} \load_h w^+_{h,o} }}%
        {\ds{r_o + \sum_{i=I+1}^{I+H} \load_h w^-_{h,o}}},\quad \forall o \in [I+H+1,N].
$$
More general feedforward networks consist of successive layers where the signals can circulate only in one direction.

\subsection{Recurrent topology}
In the case of recurrent networks circuits are allowed.
%
%
The existence of directed cycles has an important impact in the model: we can not compute the rate activities of the output neurons as functions of the network inputs (except, of course, when $N \leq 4$).
A RNN with circuits connects to the concept of dynamical systems, rather than to functions,
there is an idea of time implicit in the model.
%
%
For simplicity we assume discrete time and we avoid to use temporal notation in $\bm{\load}$.
At each time instant, the network is characterized by an internal state $\bm{\load}$ formed by the activity rates $\bm{\load}=(\load_1,\ldots,\load_N)$.
When an input pattern is presented to the network, the network updates its internal  state.
For computing the network state we must solve the system of equations~(\ref{rhos}),~(\ref{T+}) and~(\ref{T-}), where the unknown parameters are $\load_i$, ${T}^+_i$ and $T^-_i$, for all $i$.
For solving this system is necessary to perform a fixed point procedure (a summary about this computation is given in~\citep{Thimotheou10}).
The output of the network is given by the state of the output neurons.
%
%
Unlike the feedforward case, a recurrent network can use its internal states to process sequences of inputs.
As a consequence, the recurrent case is often used for solving problems where the dataset presents temporal dependencies.

\section{Random Neural Networks in supervised learning problems}
\label{SupervisedLearningTool}
In this Section we present the algorithms used for \textit{learning}.
The Section starts with a formal definition of the supervised learning problem.
Next, we present the algorithms of Gradient Descent type for training the RNN.
Then, we introduce the algorithms that use the Hessian or an approximation of the Hessian matrix for training the RNN.
We close the Section with a  general discussion that covers topics such as: limitations of the algorithms in the numerical optimisation, analysis of the algorithmic time complexity, applications of the RNN concepts in the Reservoir Computing area, a discussion about the computational power of the RNN for approximating any regular function, and an analogy of the model with other NNs.

\subsection{Specification of a supervised learning problem}
\label{ProblemDescription}
We begin by specifying a supervised learning problem.
Given a dataset~$\mathcal{L}=\{(\vak,\vbk), k=1,\ldots,K\}$, where $\vak \in \cal A$ and $\vbk \in \cal B$, with $\cal A$ and $\cal B$ some given finite dimensional spaces (typically, sets of real vectors, or of vectors of elements in some alphabet, or a mix of both types of objects).
The learning procedure consists in inferring a mapping $\nu(\va,\mathcal{L})$ in order to predict the $\vb$ values, such that some distance $d(\nu(\vak,\mathcal{L}),\vbk)$ is minimized for all $k \in \{1,2,\ldots,K\}$.
We denote by $I$ the dimension of the input vector $\va$ and $O$ the dimension of the output vector $\vb$.
For each instance $\va^{(k)}$, let us denote $\bm{\load}^{(k)}$ the output produced by the network, that is $\bm{\load}^{(k)}=\nu(\vak,\mathcal{L})$.
The distance above referred is a function $L(\cdot)$ named \textit{loss function} or \textit{cost function} that measures the deviations of the model predictions $\bm{\load}$s and the targets $\mathbf{b}$s.
Several types of loss functions have been used, the main examples are the criteria of \textit{Sum-of-Squared Errors} ($L_{\text{RSS}}$) and the \textit{Kullback-Leibler distance} ($L_{\text{KL}}$), also called \textit{cross-entropy}~\citep{HastieTibshirami,Schuma96}.
The RSS is defined as
\beq
\label{ErrorSumSquare}
L_{\text{RSS}}=\ds{\sum_{i=1}^{N}\sum_{k=1}^{K}c_i\big(b_i^{(k)}-\load_i^{(k)}\big)^2},
\eeq
where $c_i=1$ when $i$ is an output neuron, otherwise $c_i=0$.
%
%
There are several slight modifications of the previous distances, one of those is the \textit{Mean Square Error (MSE)} given by:
\beq
\label{ErrorGelenbe}
\text{MSE}=\ds{\frac{1}{K}L_{\text{RSS}}}.
\eeq
%
%
%
In supervised learning when the targets are categorical or discrete variables the problem is called \textit{classification problem}; when the target is a real vector, the problem is called  \textit{regression problem}.
%
%
%

\subsection{Random Neural Network as a learning tool}
\label{GeneralIntro}
A first approach for applying the RNN model in supervised learning tasks was introduced at the beginning of the $90$s by Erol Gelenbe~\citep{Gel93:Learning}.
This procedure is based on the classical backpropagation algorithm~\citep{RumelhartChapter8}.
%
%
As in practice, the input and output variables in learning problems are bounded with
known bounds, the algorithm described in~\citep{Gel93:Learning} assumes that $\va^{(k)}\in \left[0..1\right]^{I}$ and $\vb^{(k)}\in\left[0..1\right]^{O}$, for all sample $k$.
%
%
The RNN model as a predictor is a parametric mapping $\nu(\va, \w^+, \w^-, \mathcal{L})$, where the parameters $\w^+$ and $\w^-$ are adjusted minimizing the loss function.
In~\citep{Gel93:Learning} was considered the quadratic error presented in the expression~(\ref{ErrorGelenbe}).
The network architecture is defined with $I$ input nodes and $O$ output nodes.
There are not additional constraints regarding the network topology, that means the network can be feedforward with one or several layers, or it can be recurrent network.
We set the port of the input neurons each time that an input pattern $\vak$ is offered to the network.
%
The inputs to the positive ports are set with the input pattern: $\lambda_i^+= a_i^{(k)}$; the negative ports of input neurons are conventionally set to zero (${\lambda_i^-}=0$).
The output of the model is a vector of the activity rates produced by the output neurons.
The adjustable parameters of the mapping are the weights connections among the neurons.
We follow this Section describing the optimization algorithms that have been introduced over the last decades.

\subsection{The gradient descent optimization algorithm}
\label{GDsection}
We can now describe the gradient-based algorithm that was used so far for training the RNN model~\citep{Gel93:Learning}.
We define two set of neurons $\mathcal{I}$ and $\mathcal{O}$ that correspond to the set of input neurons and the output neurons, respectively.
The weights are initialized at some arbitrary values $\wpuvi$ and $\wnuvi$, for all $u$ and $v$.
At the $\tau$th-iteration, we select a data pattern $\big(\va^{(k)},\vb^{(k)}\big)$,~\mbox{$k=1,\ldots, K$,} where $k=\tau-1 \bmod K+1$.
The weight correction is computed following the \textit{delta learning rule}~\citep{RumelhartChapter8}, meaning that the weight correction is proportional to the partial derivative of the loss function with respect to each weight.
From~(\ref{probOutput}), the service rate of neuron~$i$ verifies
\begin{equation}
\label{rateWeights}
	r_i = \ds{ \frac{1}{1-d_i} {\ds{\sum_{j=1}^{N}
					 \Bigl( \wpij+\wnij \Bigr) } }},
\end{equation}
for all $i\in\mathcal{I}\cup\mathcal{H}$. Also note that $r_i$ is a free-parameter when $i$ is an output neuron.

At each step $\tau$, the current weight value descends in the direction of the negative gradient of $L(\cdot)$; the update rule for positive and negative weights (denoted with superscript $*$) of any connection $(u,v)$ is:
\begin{equation}
\label{deltaRule}
\wuvt= \wuvtm + \delta^{*(\tau)}_{u,v},
\end{equation}
where
\begin{align}
\label{delta}
\delta^{*(\tau)}_{u,v}
& = - \eta \sum_{i=1}^{N} c_i (\loadi^{(k)}-b^{(k)}_i)
	\left. \frac{\partial \loadi^{(k)}}{\partial w^*_{u,v}}\right|_{\w=\w^{(\tau-1)}}\nonumber\\
&= - \eta \left( \ds{\frac{\partial}{\partial \wuv}
	\frac{1}{2}\sum_{i=1}^{N} c_i\big(\loadi^{(k)} - b_i^{(k)}\big)^2}
			\right)_{\!\! \w=\w^{(\tau-1)}} \nonumber \\
&= - \eta \sum_{i=1}^{N}c_i\big(\loadi^{(k)} - b_i^{(k)}\big)
		\left. \frac{\partial \loadi^{(k)}}{\partial \wuv}
										\right|_{ \w=\w^{(\tau-1)}}\hspace{-3.5em}.			
\end{align}
The parameter~$\eta \in \left[0,1\right]$ is called \textit{learning factor}.
It is used for tuning the convergence speed of the algorithm.
%
%
Here, we set $c_i=1$ for all output neuron~$i$, otherwise $c_i=0$.
Equation~(\ref{delta}) leads to the following simplified expressions.
For each connection $(u,v)$, define the vectors~$\vg^+_{u,v}$ and~$\vg^-_{u,v}$
by
	$$\gamma^+_{uv;i}=\left\{
	\begin{array}{cl}
		\ds{-\frac{1}{r_i+T_i^-}}, & \mbox{if $u=i, \quad v\neq i$},\\
		\ds{\frac{1}{r_i+T_i^-}}, & \mbox{if $u\neq i,\quad v= i$},\\
		0, & \mbox{otherwise},
	\end{array} \right.
$$	
and
$$
	\gamma^-_{uv;i}=\left\{
	\begin{array}{cl}
	\ds{-\frac{1+\varrho_i}{r_i+T_i^-}}, & \mbox{if $u=i, \quad v = i$},\\
	\ds{-\frac{1}{r_i+T_i^-}}, & \mbox{if $u= i,\quad v \neq i$},
	\\
	\ds{-\frac{\varrho_i}{r_i+T_i^-}}, & \mbox{if $u\neq i, \quad v=i$},
	\\ 0, & \mbox{otherwise}.
	\end{array} \right.
$$
Then, denoting by $\bm{\load}$ the vector of activity rates $\bm{\load}=(\load_1,\ldots,\load_N)$:
\beq
\label{derivLoad}
	\ds{\frac{\partial \bm{\varrho}}{\partial \op _{u,v}}}
		= \bm{\gamma}^+ _{u,v} \varrho_u \left[\mathbf{I}-\bm{\Omega}\right]^{-1}
\qquad
\rm{and}
\qquad
	\ds{\frac{\partial \bm{\varrho}}{\partial \om _{u,v}}}
		= \bm{\gamma}^- _{u,v} \varrho_u \left[\mathbf{I}-\bm{\Omega}\right]^{-1},
\eeq
where $\mathbf{I}$ and $\bm{\Omega}$ are $N$-dimensional matrices, $\mathbf{I}$ is the identity,
and the element $(i,j)$ of $\bm{\Omega}$ is given by
	\beq\label{Omega}
	\Omega_{i,j} = \frac{ w^+_{i,j} - w^-_{i,j}\load_j }{ r_j+T_j^- }.
	\eeq

%

%
%

The partial derivatives were explicitly computed for a feedforward RNN with a single layer in~\citep{Kocak11}.

An \textit{online version} of the \textit{Gradient Descent (GD)} algorithm is an iterative method that processes the input patterns one-by-one realizing the following two main operations: to compute the direction of the gradient of the loss function and to update the weights using the expression~(\ref{deltaRule}).
%
The method can either be stopped using an arbitrary number of iterations or when the performance measure is smaller than some threshold value.
The online version of the GD algorithm is specified in Algorithm~\ref{AlgoGD}.
In contrast, an \textit{offline} training scheme (also called \textit{batch} algorithm) uses the whole pattern data before modifying the model parameters.
An input is offered to the network, the direction of the gradient is computed.
When all data have been presented, the gradient directions are averaged. Finally, each weight is updated using the average of the gradient directions.
In the Machine Learning literature coexists two opposite views concerning these two training schemes.
As far as we know there has been no consensus on which scheme (on-line or offline) is more efficient for training a learning model~\citep{Nakama09,WilsonMartinez03}.
%

%
\begin{algorithm}[h!t]
\caption[Specification of the GD learning algorithm for RNN (online version)]{Specification of the GD learning algorithm for the RNN model (online version).}
\label{AlgoGD}
	\SetKwInOut{Input}{Inputs}
	\SetKwInOut{Output}{Outputs}
	\Input{$\{(\va^{(k)},\vb^{(k)}):k=1,\ldots,K\}$ (training dataset),~$\eta$ (learning rate),~$\rm{maxIters}$ (max. number of iterations), the topology of the RNN (that is, the routing probabilities)}
	\Output{$\w=\{w^+_{i,j}, w^-_{i,j}:i,j=1,\ldots,N\}$ (network's weights)}
	\BlankLine
	$\tau=0$\;
	Initialize all weights (for instance, randomly);
	\tcp{we get $w_{u,v}^{*(0)}$ for all $u,v$ with $u$ either input or hidden neuron}
	Choose the value of $r_i$ for all $i \in \cal O$\;
	%
	\While{$\big((\tau<\rm{maxIters})$ or (until convergence)$\big)$}{
   		$\tau = \tau + 1$; \tcp{iteration step $\tau$}
		$k = \tau - 1 \bmod K+1$\;
		$\bm{\lambda^+}=\va^{(k)}$; \tcp{read input}

		For all $i \not\in \cal O$ compute $r_i$ using (\ref{rateWeights});
									\tcp{weights are those at $\tau-1$}
		For all $i$, compute $\load_i$
			using~(\ref{rhos}),~(\ref{T+}) and~(\ref{T-})\;
		Compute $\left[\mathbf{I}-\bm{\Omega}\right]^{-1}$ (see (\ref{Omega}))\;
		For all connections $(u,v)$, update $w_{u,v}^+$ and $w_{u,v}^-$
			using~(\ref{deltaRule}); \tcp{see also \ref{NonNegativityWeights} for
			many relevant technicalities}
		Evaluate convergence.
	}
\end{algorithm}

\subsubsection{Slight modification of the gradient descent algorithm}
\label{Contribution}
A slight variation of the GD algorithm for RNN was proposed in~\citep{BasterrechSocpar2013RNN}.
The authors increase the amount of adjustable parameters during the training of the gradient descent algorithm without modifying the network topology and the time complexity of the algorithm.
%
They consider as adjustable parameters in the training objective the following ones: the connection weights $\{\wpij,\wnij: i,j\in[1,N]\}$, the positive and negative input signals from the environment $\lambda^+_i$, $\lambda^-_i$ for all hidden and output neuron $i$, and the service rate $r_i$ for all output neuron $i$.

Considering the training error given by the expression~(\ref{ErrorGelenbe}), the update learning rule is given as follows.
Let $\bm{\Delta}$, $\bm{\Rho}$, $\bm{\Lambda^+}$ and $\bm{\Lambda^-}$ be matrices of dimensions $N\times N$, where
the matrix $\bm{\Delta}$ has elements
$$
\Delta_{i,j}=0 \text{ if }i\neq j \qquad \rm{and} \qquad \Delta_{i,i} = r_i + T_i^-,
$$
the matrix $\bm{\Rho}$ is defined as
$$
\Rho_{i,j}=0 \text{ if }i\neq j, \qquad \rm{and} \qquad    \Rho_{i,i}=\loadi,
$$
and the matrices $\bm{\Lambda}^+$ and $\bm{\Lambda}^-$ have at the position~$(i,u)$ the value~$\ds{\frac{\partial\load_i}{\partial \lambda^+_u}}$ and $\ds{\frac{\partial\load_i}{\partial \lambda^-_u}}$, respectively.
By computing the elements of $\bm{\Lambda}^+$ and $\bm{\Lambda}^-$ we have:
$$\bm{\Lambda}^+=\bm{\Delta}^{-1}((\mathbf{I}-\bm{\Omega})^{-1})^{\text{T}} \qquad {\rm{and}}
\qquad \bm{\Lambda}^-=\bm{\Delta}^{-1}\bm{\Rho} ((\mathbf{I}-\bm{\Omega})^{-1})^{\text{T}},
$$
where $\bm{\Omega}$ was defined in the expression~(\ref{Omega}).
Then, for each input pattern $(\va^{(k)},\vb^{(k)})$ at the $\tau$th iteration, we have the following update rule:

%
%
\begin{itemize}
\item[$$]
\begin{equation}
\label{PesosActualizacion}
\ds{w^{*(\tau)}_{u,v} = w^{*(\tau-1)}_{u,v}-\eta\varrho_u^{(\tau)} \bm{\gamma}^{*(\tau)}_{u,v}\left[\mathbf{I}-\bm{\Omega}\right]^{-1}(\bm{\load}^{(\tau)}-\vb^{(\tau)})},\quad \forall u,v,
\end{equation}
\item[$$]
\begin{equation}
\label{LambdaActualizacion}
\ds{\lambda_u^{*(\tau)} = \lambda^{*(\tau-1)}_{u}-\eta_1{\bm{(\load}^{(\tau)}-\vb^{(\tau)})^{\text{T}}}\Lambda^*_u},\quad \forall u\in\mathcal{H}\cup\mathcal{O},
\end{equation}
\item[$$]
\begin{equation}
\label{rhoActualizacion}
\ds{r_u^{(\tau)}=r_u^{(\tau-1)}-\eta_2\ds{\big(\load_u^{(\tau)}-b_u^{(\tau)}\big)\frac{-T_u^+}{(r_u+T_u^-)^2}}},\quad \forall u\in\mathcal{O},
\end{equation}
\end{itemize}
where $\left[\mathbf{I}-\bm{\Omega}\right]^{-1}$, $\bm{\Lambda}^*$ and $T^*$ are computed using the current input $(\va,\vb)$ and $\Lambda^*_u$ denotes the column $u$ of the matrix $\bm{\Lambda}^*$.

\subsubsection{Technical issues}
\label{NonNegativityWeights}

We discuss here some technical issues related to the learning process, well illustrated by the GD procedure.
Recall that the model can be seen as a network of queues (it is actually born in
this way).
This has some consequences, that have an impact on the design algorithmic
decisions.
A first point concerns the use of~(\ref{deltaRule}) for updating the weights.
Indeed, it may happen that~(\ref{deltaRule}) leads to a new value
for some weight that is negative or null. This does not fit the analogy with
a network of queues, or even a network of spiking neurons where the weights model
mean throughputs of spikes: weights should be positive numbers. We can
accept a null value for some $w_{u,v}^{*}$ interpreted as the fact that
there is actually no such connection between $u$ and $v$, but a negative one
has no interpretation.
The usage is to respect this analogy, modifying the updating rule such that the
weights are never negative.
Three possible approaches are proposed in~\citep{Gel93:Learning}:
\begin{itemize}
\item To use the following updating rule
\beq
\label{upRuleGD}
\wuvt= \max \left\{ \wuvtm + \delta^{*(\tau)}_{u,v},0\right\},
\eeq
and in the case that some weight is assigned value zero, then to apply one of the following rules:
\bit
\item fix a null value to this weight, and do not change it anymore in future iterations;
\item assign a zero value to this weight, but allow positive updates in subsequent iterations, keeping using~(\ref{upRuleGD}).
\eit
\item Another option is to decrease the value of $\eta$ and update again the weight using~(\ref{deltaRule}).
If the new weight is still negative, repeat until obtaining a positive number or stop  the loop using some control parameter.
Formally, this means that the learning factor becomes a variable parameter in the method.
%
In a nutshell, the global idea in descent methods
is to decrease little by little the learning factor, as we get closer and closer to a
local minimum. Global accuracy can also be improved (but also cost) if $\eta^{(\tau)}$, say,
is built by a supplementary optimization process (this is called \emph{line searching} in
the area)~\citep{numRecipes3ed}. We do not enter these details here.
\item An alternative option was presented in~\citep{Likas00}.
The authors propose a change of variable: instead of using $w_{u,v}^*$  they use new variables $\beta^+_{u,v}$ and $\beta^-_{u,v}$, such that
$$
w_{u,v}^+=\big(\beta^+_{u,v}\big)^2 \qquad \text{and} \qquad w_{u,v}^-=\big(\beta^-_{u,v}\big)^2.
$$
%
Then, instead of using the expression~(\ref{derivLoad}), we proceed as follows
\beq
	\ds \frac{\partial\load}{\beta_{u,v}^+} = 2{\beta_{u,v}^+}\frac{\partial\load}{\partial w_{u,v}^+}, \qquad
	\ds \frac{\partial\load}{\beta_{u,v}^-} = 2{\beta_{u,v}^-}\frac{\partial\load}{\partial w_{u,v}^-}.
\eeq
\end{itemize}

\subsubsection{Computational cost of the gradient descent algorithm}
%
When one data pattern is presented to update each weight in the network the main computational effort consists of computing  $\left[\mathbf{I}-\bm{\Omega}\right]^{-1}$ using~(\ref{derivLoad})~\citep{Gel93:Learning}.
This effort has $O(N^3)$ time complexity.
A remark made in~\citep{Gel93:Learning} consists in that when a $m$-step relaxation method is applied the time complexity decreases to $O(mN)$.

Additionally, the general scheme of the algorithm can be adapted when we use a feedforward RNN.
In this case the matrix $\mathbf{I}-\bm{\Omega}$ becomes triangular, so the computational cost of computing its inverse decreases to $O(N^2)$.
Also, the computational effort to compute each activity rate in feedforward networks is reduced, due to the the activity rate of any neuron depends only on the neurons in the preceding layers.

\subsection{Second order optimization methods}
\label{QN}
%
In this Section, we present the optimisation methods for RNN that use the information given by the second derivative of the loss function. We start introducing the
\textit{Gauss-Newton (GN)} methods, next we explore the  \textit{Quasi-Newton (QN)} techniques.
We present four particular algorithms developed for training RNNs: the \textit{Broyden-Fletcher-Goldfarb-Shanno (BFGS)}, the \textit{Davidon, Fletcher and Powell (DFP)}, the \textit{Levenberg-Marquardt (LM)} and the \textit{LM with Adaptative Momentum (LM-AM)}.

The \textit{Gauss-Newton (GN)} algorithm is a technique for solving non-linear least squares problems that incorporates the second derivatives of the loss function or an approximation of those.
Unlike the algorithms of first derivatives that can solve a large non-sparse optimization problems, a GN method can only be used when the loss function is given by a quadratic objective function, for instance the expression~(\ref{ErrorGelenbe}).
The methods of the GN type are generally considered more powerful in terms of accuracy and time than the algorithms that only use the first derivative information.

The GN method is based on an expansion of the loss function in the Taylor series.
Let $M$ be the number of adjustable parameters (the number of weights $\wpij$ and $\wnij$).
We define the $M$-dimensional vector $\mathbf{w}$ that collects in some arbitrary order the weights $\wpij$ and $\wnij$.
Let $\va$ be an input vector on the network.
The GN algorithm employs a linear approximation with the first three terms of the Taylor series
\begin{equation}
\label{TaylorGN}
%
%
L(\va,\w+\bm{\delta}) \approx L(\va,\w)+\ds{\sum_{m=1}^{M}\frac{\partial L(\va,\w)}{\partial w_m} \delta_m+ \sum_{i,j}^{M}\frac{\partial L(\va,\w)}{\partial w_i\partial w_j}\delta_i\delta_j},
\end{equation}
where $\bm{\delta}$ is a $M$-dimensional vector that represents a small correction of the weights.
The solution is found by solving the $M\times M$ set of equations (called \textit{normal equations})
\begin{equation}
\label{normalEquationGN}
\mathbf{J}^{\text{T}} \mathbf{J}\bm{\delta}=-\mathbf{G},
\end{equation}
where $\mathbf{G}$ and $\mathbf{J}$ are the gradient vector and the Jacobian matrix, respectively.
For computing $\mathbf{G}$ and $\mathbf{J}$ we proceed as follows.
%
Let $\mathbf{e}^{(k)}$ be the \textit{residual} row vector of dimension $\mathcal{O}$ for the $k$th input-output training pair,
\begin{equation}
\label{residual}
\mathbf{e}^{(k)} = \vb^{(k)}-\bm{\load}^{(k)}.
\end{equation}
Collecting those residuals, we have a vector $\mathbf{E}$ of $S\times 1$ dimensions, with $S=K\mathcal{O}$.
Then, the gradient vector of $L(\cdot)$ has $M\times 1$ dimensions and its $m$th element is
\beq
\label{gradient}
G_m=\ds{\sum_{s=1}^{S}\frac{\partial e_s}{\partial w_m}e_s}.
\eeq
The Jacobian matrix has dimensions $S\times M$ and its $(s,m)$ element is
\newcommand{\myGrad}{\triangledown}
\begin{equation}
\label{Jacobian}
{J}_{s,m} = \ds{\partial E_s/\partial w_m} .
\end{equation}
For computing the partial derivatives of~(\ref{gradient}) and~(\ref{Jacobian}) we use the expressions presented in~(\ref{derivLoad}).

The GN method is a batch type algorithm. We call an \textit{epoch} of the GN algorithm when all the patterns in the training set are used~\citep{Bengio2000}.
At each epoch $\tau$, the weight correction $\delta$ is computed, next the weights are updated as follows:
\begin{equation}
\label{weightUpdateGN}
\w^{(\tau)}=\w^{(\tau-1)}+\alpha^{(\tau)}\bm{\delta}^{(\tau)},
\end{equation}
where $\alpha\in(0,1]$ is computed using a line search technique~\citep{numericalRecips92}. In the canonical GN method this parameter is set to~$1$.
A better strategy is tuning $\alpha$ with less values until some suitable point. For details about how to tune $\alpha$ see Chapter 9 of~\citep{numericalRecips92}.

The GN method for solving the problem of minimization using NNs presents several drawbacks.
The method requires a good initial solution, that is often not available~\citep{LeCun92}.
Another drawback is that the GN method requires computing the Hessian matrix $\mathbf{H}$ ($\mathbf{H}=\mathbf{J}^{\text{T}} \mathbf{J}$) and its inverse, both computations can be expensive.
Therefore, the method is expensive in time and in storage.

A \textit{Quasi-Newton (QN)} method type is a variant of the GN algorithms that uses an approximation of the Hessian matrix ($\widetilde{\mathbf{H}}$) for solving the normal equations.
The general approach behind a QN method is an iterative procedure that consists of starting with a positive and symmetric matrix and updating it in successive steps in such a way that the matrix remains positive definite and symmetric.
The update rule always moves in a downhill direction for solving the normal equations and guarantees that $\widetilde{\mathbf{H}}$ approximates $\mathbf{H}$.
As we already commented so far, the implementation of the second order methods is offline, thus at each epoch the network outputs are computed for the whole of input patterns.
We present in Schema~\ref{AlgoSchema} a procedure that shows how to compute those model outputs.
In the following of this Section we will use this schema as a black box being a part of the GN and Quasi-Newton algorithms.
In the remainder of this Section, we present four algorithms based on approximations of the Hessian matrix.

\begin{algorithm}[h!t]
\caption{Auxiliary schema. Given a RNN the procedure shows how to compute the network outputs for the whole input dataset. The procedure returns a $K\times N$ matrix, that has the vector $\bm{\load^{(k)}}$ computed with the input pattern $\vak$ in its $k$-row.}
\label{AlgoSchema}
	\SetKwInOut{Input}{Inputs}
	\SetKwInOut{Output}{Outputs}
	\Input{$\{(\va^{(k)},\vb^{(k)}):k=1\ldots,K\}$ (training dataset),
	the topology of the RNN}
	\Output{The neuron activity rate produced by the whole of input patterns:  $C$ a $K\times N$ matrix}
	\BlankLine
	Choose the value of $r_i$  for all output neuron $i$\;
	For all $i\notin \mathcal{O}$ compute $r_i$ using~(\ref{rateWeights})\;
	\For{$(k\leftarrow 1$ \KwTo $K)$}{
				$\bm{\lambda}^+=\vak$;
					\tcp{read input}
				For all $i$, compute $\load_i^{(k)}$ using~(\ref{rhos}),~(\ref{T+}) and~(\ref{T-})\;
				\tcp{see also \ref{NonNegativityWeights} for
			many relevant technicalities}
				Set the row $k$ of $C$ with the vector $\bm{\load}^{(k)}$\;
	}
\end{algorithm}

\subsubsection{The Broyden-Fletcher-Goldfarb-Shanno algorithm}
\label{BFGSAsubsection}
\newcommand{\deltaW}{(\w^{(\tau)}-\w^{(\tau-1)})}
\newcommand{\deltaG}{(\mathbf{G}^{(\tau)}-\mathbf{G}^{(\tau-1)})}
The \textit{Broyden-Fletcher-Goldfarb-Shanno (BFGS)} method for the RNN model was introduced in~\citep{Likas00}.
%
%
The BFGS is an offline algorithm, which at each epoch $\tau$ an approximation of the Hessian matrix ${\widetilde{\mathbf{H}}}^{(\tau)}$ is computed.
The method starts using the identity matrix as the initial Hessian approximation ${\widetilde{\mathbf{H}}}^{(0)}=\mathbf{I}$.
The Choleski factorization is used for decomposing a symmetric and positive definite matrix into two triangular matrices.
Choleski factorization is more efficient than alternative methods for solving linear equations, it is about two times faster than the alternative ones.
For details about the implementation of this factorization see~\citep{numericalRecips92}.
The matrix ${\widetilde{\mathbf{H}}}^{(\tau)}$ is decomposed using Choleski factorization as
\begin{equation}
\label{L_BFGS}
{\widetilde{\mathbf{H}}}^{(\tau)}=\mathbf{L}^{(\tau)}{\mathbf{L}^{\text{T}(\tau)}}.
\end{equation}
Let $c$ be an auxiliary scalar defined at each epoch as
\begin{equation}
\label{c_BFGS}
{c^{(\tau)}}^2=\ds{\frac{\deltaW^{\text{T}}\deltaG}{\deltaW^{\text{T}}{\widetilde{\mathbf{H}}}^{(\tau)}\deltaW}}.
\end{equation}
We define an auxiliary vector $v$ as
\begin{equation}
\label{v_BFGS}
\mathbf{v}^{(\tau)}=c^{(\tau)}\mathbf{L}^{(\tau)}\deltaW.
\end{equation}
Next, we compute
\begin{equation}
\label{A_BFGS}
\mathbf{A}^{(\tau)}=\mathbf{L}^{(\tau)}+\ds{\frac{(\deltaG-\mathbf{L}^{(\tau)}\mathbf{v}^{(\tau)}){\mathbf{v}^{\text{T}(\tau)}}}{{\mathbf{v}^{\text{T}(\tau)}}\mathbf{v}^{(\tau)}}}.
\end{equation}
The update of the Hessian matrix approximation is given by
\begin{equation}
\label{H_BFGS}
{\widetilde{\mathbf{H}}}^{(\tau+1)}=\mathbf{A}^{(\tau)}{\mathbf{A}^{\text{T}(\tau)}}.
\end{equation}
Finally, the weight update is given by $\bm{\delta}$ solving
\begin{equation}
\label{delta_BFGS}
{\widetilde{\mathbf{H}}}^{(\tau+1)}\bm{\delta}^{(\tau+1)}=-\mathbf{G}^{(\tau)}.
\end{equation}

In summary, the BFGS method for RNN presented in~\citep{Likas00} is defined in Algorithm~\ref{AlgoBFGS}.
%
%
%
\begin{algorithm}[h!t]
\caption[Specification of the BFGS algorithm for the RNN model]{Specification of the BFGS algorithm for the RNN model.}
\label{AlgoBFGS}
	\SetKwInOut{Input}{Inputs}
	\SetKwInOut{Output}{Outputs}
	\Input{$\{(\va^{(k)},\vb^{(k)}):k=1\ldots,K\}$ (training dataset),
	maxIters (max. number of iterations), the topology of the RNN
	}
	\Output{The weights: $\w=\{w^+_{i,j}, w^-_{i,j}:i,j=1,\ldots,N\}$ (network's weights)}
	\BlankLine
	$\tau=0$\;
	Initialize all weights (for instance, randomly);
	\tcp{we get ${\w^*_{u,v}}^{(0)}$ for all $u$, $v$ with $u$ either input or hidden neuron}
	Choose the value of $r_i$  for all output neuron $i$\;
	${\widetilde{\mathbf{H}}}=\mathbf{I}$\;
	Compute $\mathbf{G}$ using~(\ref{gradient}) and~(\ref{derivLoad})\;
	Compute~$\bm{\delta}$ solving~(\ref{delta_BFGS})\;
	For all connections $(u,v)$, update $w_{u,v}^+$ and $w_{u,v}^-$ using~(\ref{weightUpdateGN})\;
%
	\While{$\big((\tau<\rm{maxIters})$ or (until convergence)$\big)$}{
	$\tau=\tau+1$\;
	Apply Schema~\ref{AlgoSchema}\;
	%
	Compute $\mathbf{G}$ using~(\ref{gradient}) and~(\ref{derivLoad})\;
	\label{CholeskiFac} Compute $\mathbf{L}$ using Choleski factorization see~(\ref{L_BFGS})\; 
	\label{terms_BFGS} Compute $c$, $\mathbf{v}$ and $\mathbf{A}$ using~(\ref{c_BFGS}),~(\ref{v_BFGS}) and~(\ref{A_BFGS}), respectively\;
	Update~${\widetilde{\mathbf{H}}}$ using~(\ref{H_BFGS})\;
	Compute~$\bm{\delta}$ solving~(\ref{delta_BFGS})\;
	For all connections $(u,v)$, update $w_{u,v}^+$ and $w_{u,v}^-$ using~(\ref{weightUpdateGN});
	\tcp{see also \ref{NonNegativityWeights} for
			many relevant technicalities
		}
	%
	Evaluate convergence\;
	}
\end{algorithm}

\subsubsection{The Davidon-Fletcher-Powell algorithm}
\label{DFPAsubsection}
The \textit{Davidon-Fletcher-Powell (DFP)} algorithm is another widely used QN method sometimes referred as Fletcher-Powell~\citep{numericalRecips92}.
%
%
The algorithm is a slight variation of BFGS algorithm, the difference between them is given in the following terms.
The scalar $c$ is defined as
\begin{equation}
\label{c_DFS}
{c^{(\tau)}}^2=\ds{\frac{\deltaW^{\text{T}}\deltaG}{\deltaG^{\text{T}}{\widetilde{\mathbf{H}}}^{(\tau)}\deltaG}},
\end{equation}
and the vector $\mathbf{v}$ is such that solves the linear system,
\begin{equation}
\label{L_DFS}
\mathbf{L}^{(\tau)}\mathbf{v}^{(\tau)}=c^{(\tau)}\deltaG.
\end{equation}
The matrix $\mathbf{A}$ is determined by computing
\begin{equation}
\label{A_DFS}
\mathbf{A}^{(\tau)}=\mathbf{L}^{(\tau)}-\ds{ \frac{\deltaG(\deltaW^{\text{T}}\mathbf{L}^{(\tau)}-{\mathbf{v}^{{\text{T}}(\tau)}})}{\deltaW^{\text{T}}\deltaG}}.
\end{equation}
Finally, yielding the Hessian approximation
\begin{equation}
\label{H_DFS}
{\widetilde{\mathbf{H}}}^{(\tau)}=\mathbf{A}^{(\tau)}{\mathbf{A}^{\text{T}(\tau)}},
\end{equation}
and we compute the search direction $\bm{\delta}$ for update the weights solving the expression~(\ref{delta_BFGS}).

According empirical results the BFGS performs better than the DFP method~\citep{numericalRecips92}.
Although, for some specific benchmark problems the DFP reached better accuracy than DFGS~\citep{Likas00}.
The algorithm is summarized in~\ref{AlgoDFS}.
\begin{algorithm}[h!t]
\caption[Specification of the DFS algorithm for the RNN model]{Specification of the DFS algorithm for the RNN model. The DFS and the BFGS algorithms differ only in details. As a consequence, we introduce the DFS referencing the schema already presented in Algorithm~\ref{AlgoBFGS}.}
\label{AlgoDFS}
	\SetKwInOut{Input}{Inputs}
	\SetKwInOut{Output}{Outputs}
	\Input{$\{(\va^{(k)},\vb^{(k)}):k=1\ldots,K\}$ (training dataset),
	maxIters (max. number of iterations), the topology of the RNN
	}
	\Output{The weights: $\w=\{w^+_{i,j}, w^-_{i,j}:i,j=1,\ldots,N\}$ (network's weights)}
	\BlankLine
	\LinesNumberedHidden
	
	\tcp{Perform the lines $1$ until $7$ of Algorithm~\ref{AlgoBFGS}.}
	\While{$\big((\tau<\rm{maxIters})$ or (until convergence)$\big)$}{
%
	$\tau=\tau+1$\;
	Apply Schema~\ref{AlgoSchema}\;
	%
	Compute $\mathbf{G}$ using~(\ref{gradient}) and~(\ref{derivLoad})\;
	Compute $c$ using~(\ref{c_DFS})\;
	Compute $\mathbf{L}$ solving~(\ref{L_DFS})\;
	Compute $\mathbf{A}$ using~(\ref{A_DFS})\;
%
%
	\tcp{Perform the lines $14$ until $17$ of Algorithm~\ref{AlgoBFGS}.}
}
\end{algorithm}

\subsubsection{The Levenberg-Marquardt algorithm}
\label{LMsubsection}
The \textit{Levenberg-Marquardt (LM)} algorithm  is one of the most standard optimization methods used in the NN area~\citep{numericalRecips92,Ampazis00,Hagan94}.
The LM is a sort of compromise between an offline version of the GD algorithm and a GN method~\citep{Marquardt63,numericalRecips92}.
The algorithm was introduced for training RNN in~\citep{Baster09}.

At each epoch $\tau$, the approximation of the Hessian matrix is given by,
\beq
\label{approxHessian}
{\widetilde{\mathbf{H}}}^{(\tau)}=\mathbf{J}^{\text{T}(\tau)}\mathbf{J}^{(\tau)}+\mu^{(\tau)}\mathbf{I},
\eeq
where $\mu^{(\tau)}>0$ is called \textit{dumping} term, $\mathbf{I}$ is the identity matrix of dimension $M\times M$, and  $\mathbf{J}$ is the Jacobian matrix that is computed using~(\ref{Jacobian}).
The dumping term $\mu$ is modified at each epoch.
In the case that the prediction error decreases, then the dumping term is reduced by some constant value $\beta$
\begin{equation}
\label{decrease}
\mu  \leftarrow \mu/\beta.
\end{equation}
Otherwise, the dumping value is increased by a factor of $\beta$,
\begin{equation}
\label{increase}
\mu  \leftarrow \mu\beta.
\end{equation}
%
So far, the factor for modifying the dumping term was set as $\beta=10$~\citep{numericalRecips92,Baster09}.

The LM algorithm computes the weight correction $\bm{\delta}$ solving the system~(\ref{delta_BFGS}).
Then, the update rule for the weights is given by the expression~(\ref{weightUpdateGN}).
In~\citep{Baster09}, this weight update considers only the search direction $\bm{\delta}$. In other words, the authors set $\alpha=1$ in the expression~(\ref{weightUpdateGN}).
The algorithm can evolve through either of extreme possible situations are~\citep{Hagan94,numericalRecips92}:
\bit
\item If the dumping term approaches to zero, the LM basically performs as the Gauss-Newton method.
\item Otherwise, when the dumping term is very large, the matrix ${\widetilde{\mathbf{H}}}$ becomes diagonal dominant, so
the update rule is similar to the updating expression of gradient descent method using a learning factor of $1/\mu$.
\eit

Concerning the stopping conditions, the method can fail if the Jacobian matrix becomes singular or nearly to singular.
Even if this situation is rare in practice, a control of the condition number of $\mathbf{J}$ can be useful~\citep{numericalRecips92}.
Besides, it is necessary to control that the dumping factor satisfies some boundary conditions.
%
%
It is not recommended to stop after an epoch wherein the training objective error increases.
For more technical discussion about the stopping criteria of the LM see~\citep{numericalRecips92}.
We present the LM procedure in Algorithm~\ref{AlgoLM}.

\begin{algorithm}[h!t]
\caption[Specification of the LM algorithm for RNN]{Specification of the LM algorithm for RNN.}
\label{AlgoLM}
	\SetKwInOut{Input}{Inputs}
	\SetKwInOut{Output}{Outputs}
	\Input{$\{(\va^{(k)},\vb^{(k)}):k=1\ldots,K\}$ (training dataset),
	maxIters (max. number of iterations), the topology of the RNN, $\mu$ (dumping term), $\beta$ (constant to modify $\mu$)
	}
	\Output{The weights: $\w=\{w^+_{i,j}, w^-_{i,j}:i,j=1,\ldots,N\}$ (network's weights)}
	\BlankLine
	%
	
	$\tau=0$\;
	Initialize all weights (for instance, randomly);
	\tcp{we get ${\w^*_{u,v}}^{(0)}$ for all $u$, $v$ with $u$ either input or hidden neuron}
	Choose the value of $r_i$  for all output neuron $i$\;
	
	\While{$\big((\tau<\rm{maxIters})$ or (until stopping conditions)$\big)$}{
	$\tau=\tau+1$\;
	Apply Schema~\ref{AlgoSchema}\;
	Compute $L(\w)$ using~(\ref{ErrorGelenbe})\;
	Compute $\mathbf{G}$ using~(\ref{gradient}) and~(\ref{derivLoad})\;
	Compute $\mathbf{J}$ using~(\ref{Jacobian})\;
	Compute ${\widetilde{\mathbf{H}}}$ using~(\ref{approxHessian})\;
	Compute $\bm{\delta}$ solving~(\ref{delta_BFGS})\;
	Compute temporal weights $\mathbf{w}_{\rm{tmp}}^{*} =\mathbf{w}^{*}+\bm{\delta}$\;
	\tcp{weights $\mathbf{w}^{*}$ are those at $\tau-1$, see also \ref{NonNegativityWeights} for technicalities}
	Apply Schema~\ref{AlgoSchema} using the weights $\mathbf{w}_{\rm{tmp}}^{*}$\;
	Compute $L(\w_{\rm{tmp}}^*)$ using~(\ref{ErrorGelenbe})\;
		\If{$\big(L(\w_{\rm{tmp}})<L(\w)\big)$}
				{
				Update $\mu$ using~(\ref{decrease})\;
				Set the weights $\w^*$ with  $\w_{\rm{tmp}}^*$\;
				Set $L(\w^*)$ with $L(\w_{\rm{tmp}}^*)$\;
		}
		\Else{
				Update $\mu$ using~(\ref{increase})\;
		}
		Evaluate stopping conditions\;
}
\end{algorithm}

\subsubsection{Levenberg-Marquardt with adaptive momentum training}
\label{SectionLM-AM}
A variation of the LM method applied to NNs was developed in~\citep{Ampazis00,Ampazis99}.
This approach was adapted for the case of RNN on learning problems in~\citep{Baster09}.
The idea consists in inserting a \textit{momentum term} that controls the directions followed in the searching space.
The algorithm was introduced under the name of \textit{Levenberg-Marquardt with Adaptative Momentum (LM-AM)}~\citep{Ampazis00,Baster09}.
The approach consists in maintaining the \textit{conjugacy} of successive searching vectors~\citep{Barnard92}.
This means that at an epoch $\tau$ the new direction ${\bm{\delta}}^{(\tau)}$ depends on the selected direction at the previous epoch ${\bm{\delta}}^{(\tau-1)}$.
It is desirable that the motion along a direction at the current step positively interferes with the minimization along the previous step.
Formally, this property occurs when both vectors ${\bm{\delta}}^{(\tau-1)}$ and ${\bm{\delta}}^{(\tau)}$ are \textit{mutually conjugate}, the same principle is used in the Conjugate Gradient algorithm~\citep{numericalRecips92}.
\newcommand{\Ht}{\widetilde{\mathbf{H}}}
\newcommand{\bg}{\mathbf{g}}
\newcommand{\HH}{\Ht^{(\tau)}}
\newcommand{\dd}{\bm{\delta}^{(\tau-1)}}
\newcommand{\gt}{\bg^{(\tau)}}
\newcommand{\frA}{\frac{\lambda_1}{2\lambda_2}}
\newcommand{\frB}{\frac{1}{2\lambda_2}}
\newcommand{\HHi}{\left[\HH\right]^{-1}}

In the LM-AM the update rule for the weights at the epoch $\tau$ is given by:
\beq
\label{deltaLM-AM}
\bm{\delta}^{(\tau)} = - \frac{\lambda_1}{2\lambda_2}\left[{\widetilde{\mathbf{H}}}^{(\tau)}\right]^{-1}\mathbf{\bm{G}^{(\tau)}}+\frac{1}{2\lambda_2} \bm{\delta}^{(\tau-1)},
\eeq
where
\beq
\label{lambda1LM-AM}
\lambda_1=\frac{-2\lambda_2\Delta Q^{(\tau)} +c_2}{c_1}
\end{equation}
\rm{and}
\begin{equation}
\label{lambda2LM-AM}
\lambda_2=\textstyle \displaystyle  \frac{1}{2} \sqrt{
\frac{c_1c_3-{c_2}^2}{c_1(\Delta P)^2 -(\Delta Q^{(\tau)})^2}
},
\eeq
with
$$\label{eq:am25}
		\Delta Q^{(\tau)}=\frac{1}{2\lambda_2} (c_2 -\lambda_1 c_1),
$$
and the constants $c_1,c_2$ and $c_3$ are three real numbers defined as follows:
\begin{equation}
\label{c1LM-AM}
c_1={\mathbf{G}^{\text{T}(\tau)}} \left[{\widetilde{\mathbf{H}}}^{(\tau)}\right]^{-1}\mathbf{G}^{(\tau)},
\end{equation}
\begin{equation}
\label{c2LM-AM}
\quad c_2 ={\mathbf{G}^{\text{T}(\tau)}} {\bm{\delta}}^{(\tau-1)}
\end{equation}
and
\begin{equation}
\label{c3LM-AM}
c_3={\bm{\delta}^{\text{T}(\tau-1)}}{\widetilde{\mathbf{H}}}^{(\tau)}{\bm{\delta}}^{(\tau-1)}.
\end{equation}
In practice, it is suggested to set~$\Delta Q$ as
\begin{equation}
\label{DeltaQLM-AM}
\Delta Q^{(\tau)}=-\zeta \Delta P \sqrt{c_1},
\end{equation}
where $\zeta$ is some constant between $0$ and $1$~\citep{Ampazis00,Baster09}.
Then, the parameters for the LM-AM procedure are $\zeta$ and $\Delta P$.
When the LM-AM is applied for optimizing classic NNs is suggested to experiment with $0.85 \leq \zeta \leq 0.95$~\citep{Ampazis00}.
In~\citep{Baster09}, the authors applied LM-AM for optimizing the RNN model achieving the best results when $0.1 \leq \Delta P \leq 0.6$.
The Algorithm~\ref{AlgoLM-AM} presents the pseudo-code of the LM-AM.

\begin{algorithm}[h!t]
\caption[Specification of the LM-AM algorithm for RNN]{Specification of the LM-AM algorithm for RNN.
This algorithm is a variation of Algorithm~\ref{AlgoLM}, as a consequence we introduce it
referencing the schema already presented there.}
\label{AlgoLM-AM}
	\SetKwInOut{Input}{Inputs}
	\SetKwInOut{Output}{Outputs}
	\Input{$\{(\va^{(k)},\vb^{(k)}):k=1\ldots,K\}$ (training dataset), maxIters (max. number of iterations), the topology of the RNN, $\mu$ (dumping term), $\beta$ (constant to modify $\mu$), $\zeta$ and $\Delta P$ (specific parameters of the LM-AM)}
	\Output{The weights: $\w=\{w^+_{i,j}, w^-_{i,j}:i,j=1,\ldots,N\}$ (network's weights)}
	\BlankLine
	\tcp{Performs the lines 1 until 3 of Algorithm~\ref{AlgoLM}}
	\While{$\big((\tau<\rm{maxIters})$ or (until stopping conditions)$\big)$}{
	\tcp{Performs the lines 5 until 10 of Algorithm~\ref{AlgoLM}}
	Compute $c_1$, $c_2$ and $c_3$ using~(\ref{c1LM-AM}), (\ref{c2LM-AM}) and~(\ref{c3LM-AM})\;
	Compute $\Delta Q$ using~(\ref{DeltaQLM-AM})\;
	Compute $\lambda_1$ and $\lambda_2$ using~(\ref{lambda1LM-AM}), (\ref{lambda2LM-AM})\;
	Compute $\bm{\delta}$ using~(\ref{deltaLM-AM})\;
	\tcp{Performs the lines 12 until 21 of Algorithm~\ref{AlgoLM}}
}
\end{algorithm}

\section{Critical review}
\label{CriticalReview}

In this section we discuss some stability issues of the model when applied to supervised learning tasks. Next, we analyze the time complexity and memory stockage of Gauss-Newton methods. We discuss the difficulties of training recurrent topologies and we present an alternative for using recurrent networks without the drawback of learning the network parameters.
Next, we present some properties of the RNN model and its analogy with a specific type of NN. The section ends with a presentation of RNN variations.
\subsection{Stability issues}
Originally, the model was introduced as a network of queues.
Some of the consequences of that were already discussed in Sec.~\ref{NonNegativityWeights}.
Actually, the model has been applied respecting the analogy with queueing networks.
As a consequence, the weights are controlled in order to keep them in positive intervals.
%
%
If we see neuron~$i$ as a queue, the interpretation of $\loadi$ makes basically sense in the stable case only, when the queue is in equilibrium (that is, when the underlying stochastic process is ergodic). The same holds for the whole network. This is a tricky point. Seeing~$i$ as a queue, we have stability only when $T^+_i < r_i + T^-_i$. If we want to
keep this true, we need supplementary constraints in the optimization processes,
since, at each step, a new neuron rate $r_i$ is computed, as a function of the
previously computed weights, and intuitively, it must be high enough such that
the neuron remains stable.
The point is also relevant regarding the use of the model in supervised learning, because roughly speaking, stability implies that the non-linear system of equations~(\ref{rhos}),~(\ref{T+}) and~(\ref{T-}) has an unique solution.
The usage is, however, to ignore this point and only check stability for the output neurons.
In this case, if for some neuron $i$ we obtain $\load_i > 1$, we replace this number by~$1$.

These remarks mean that there is an open research line here, where other learning
schemes could be designed weakening the connexion with the queuing world.

\subsection{Time complexity}
The LM algorithm has been proved to be efficient in terms of computational time and accuracy rate.
In spite of that, the method presents some drawbacks.
One of the weak points is that it may require a large amount of memory, due to the need of storing large matrices~\citep{Barnard92}.
The procedure is offline, at each iteration the algorithm uses the whole data pattern for computing the Jacobian matrix and the inverse of the pseudo-Hessian matrix.
These two operations can be expensive when the dimensions of both matrices are large.
An exact and efficient method for computing the Hessian for the feedforward case was introduced in~\citep{Bishop92}.
However, this approach is not practical in the case of large networks~\citep{Martens12ConfICML}.
Another operation that has a high computational cost is the computation of the inverse of $\widetilde{\mathbf{H}}$.
For a $M\times M$ matrix, we have the following well known techniques for inverting a matrix and their corresponding computational costs: Gauss-Jordan elimination method (time complexity ${O}(M^3)$)~\citep{numericalRecips92}, Strassen method (time complexity ${O}(M^{\log_2 7 }$))~\citep{numRecipes3ed}, Coppersmith-Winograd method (time complexity ${O}(M^{2.376})$)~\citep{Coppersmith90}.
A slight variation of the LM algorithm was proposed in~\citep{Ukil07}, wherein the authors reduce the memory space used for storing the Jacobian and pseudo-Hessian matrix.
However, the proposal is much slower in terms of convergence speed.
%

\subsection{Difficulties for optimizing recurrent topologies}
In the machine learning community there have been numerous efforts to develop algorithms for training a NN with a recurrent topology.
In spite of that, in practice is hard to train recurrent networks.
An algorithm based on the gradient information has often stability problems, due to the volatile relationship between the weights and the states of the neurons~\citep{Martens11}.
These phenomena were studied by several researchers.
In the literature, they are identified as \textit{vanishing} and the \textit{exploding} gradient problems~\citep{Bengio94}.
The first one occurs when the gradient norm tends fast towards zero.
The exploding gradient phenomenon refers to the opposite situation, that is, when the gradient norm tends to get very large~\citep{Pascanu13}.
As far as we know, the vanishing and exploding gradient phenomena have not been yet studied  for RNNs.
The research effort continues to address these issues. For instance, a new attempt to train NN with recurrences was recently introduced under the name of \textit{Hessian-Free Optimization}~\citep{Martens11}.
This could also be explored for training recurrent RNNs.

\subsection{Reservoir computing and Random Neural Networks}
During the last fifteen years, \textit{Echo State Network (ESN)} has received much attention in the NN community, due to its good performance for solving time-series learning problems.
The ESN introduces a new approach to design and train NN with recurrences.
In these models, learning only occurs in the weights that are not involved in recurrences.
Those involved in circuits are deemed fixed during the training process.

In the canonical ESN the neurons have associated a sigmoid transfer function.
A variation of the ESN model named \textit{Echo State Queueing Network (ESQN)} that uses the dynamics of the RNN (based on Eqns.~(\ref{rhos}), (\ref{T+}) and (\ref{T-})) was introduced in~\citep{Baster12ESQN}\citep{BasterACM13}.
The ESQN has been successfully applied for solving temporal learning tasks, for instance, in predicting future Internet traffic based on past observations~\citep{Baster12ESQN}.

\subsection{Universal approximator}
George Cybenko investigated in~$1989$ the conditions under which feedforward NNs are dense functions in the space of continuous functions defined, say, in the hypercube $[0,1]^{\left|\mathcal{O}\right|}$)~\citep{Cybenko89}.
Cybenko proved that any continuous function can be uniformly approximated by a continuous classic NN having a finite number of neurons and only one hidden layer.
The considered activation function of the neurons was a sigmoid function.
In~\citep{GelMao99}, the authors studied this property for the RNN model.
They proved that RNNs constitute a family of functions that can approximate any real continuous function~$f:[0,1]^{\left|\mathcal{I}\right|}\rightarrow [0,1]^{\left|\mathcal{O}\right|}$ with an arbitrary precision~\citep{GelMao99,Gelenbe2004b}. 
In order to prove that RNNs satisfy this \textit{approximator universal property}, the authors consider a specific topology and two extensions of the model called \textit{Bipolar Random Neural Network (BRNN)} and \textit{Clamped Random Neural Network (CRNN)}.
For details about the proof see~\citep{Gelenbe1991d,Gel98:LearningAndApprox,GelMao99,Gelenbe2004b}.

\newcommand{\ip}{u}
\newcommand{\jp}{v}

\subsection{Analogy between RNNs and other NNs}
In~\citep{Gel89:RNNPosNeg} was studied an analogy between a specific class of Artificial NN and the RNN model.
%
%
Consider a classical NN and let $w_{uv}$ be the weight from neuron $u$ to neuron $v$ (note that the notation is in reverse order to that used in the standard NN literature~\citep{Bishop95,RumelhartChapter2}).
Given a sequence of inputs $y_1,y_2, \ldots, y_N$ to neuron $\ip$, its output is
\begin{equation}
y_{\ip}=f\bigg(\ds{\sum_{\jp=1}^{N}w_{\jp\ip} y_{\jp} - \theta_{\ip}}\bigg),
\end{equation}
where $f(\cdot)$ is a sigmoid function~\citep{Gel89:RNNPosNeg}.
%
Assume that the input-output pattern in the training data is binary.
The input space is $\{0,1\}^{\left|\mathcal{I}\right|}$ and the output space is $\{0,1\}^{\left|\mathcal{O}\right|}$, where ${\left|\mathcal{I}\right|}$ and ${\left|\mathcal{O}\right|}$ are positive integers.
The network topology is of the feeedforward type with multiple layers.
The adjustable parameters of the model are the weight connections (${w}_{\ip\jp}$) and the bias parameter ($\theta_{\ip}$) for $\ip,\jp=1,\ldots,N$.
Let us consider the indexes $\ip$ and $\jp$ to denote the neurons in the ANN and the indexes $i$ and $j$ to denote the neurons in the RNN model.
The analogy between both networks is built using the same number of input, hidden and output neurons in the two networks, as follows.
%
%
%
The threshold of neuron $\ip$ is associated with the flow of negative signals to neuron $i$: $T_i^-=\theta_{\ip}$.
The relation among the weight connections is given by the following rules:
if $w_{\ip\jp}>0$, then $r_i\ppij=w_{\ip\jp}$; if $w_{\ip\jp}<0$, then $r_i\pnij=|w_{\ip\jp}|$.
If $i$ is an output neuron, then $d_i=1$, and $r_i$ is a free parameter.
The authors propose to set $d_i=0$ and $r_i=\sum_{j=1}^{N} \bigl( \ppij+\pnij \bigr) = \sum_{\jp=1}^{N}|w_{\ip\jp}|$, when $i$ and $\ip$ are an input and a hidden neuron respectively.
Then, the input signals of an input node of the RNN are set as follows:
\bit
\item If the element $i$ of the input pattern is equal to $1$, then $\lpi$ is a non null constant in $(0,1]$ while $\lnni=0$.
The authors propose an ad-hoc setting for the input rate $\lpi$, chosen according to the training data in order to obtain the desired outputs.
%
\item If the input pattern is equal to $0$, then $\lnni\neq 0$ and $\lpi=0$.
\eit
The vector of output activation values $(y_1,\ldots,y_{N})$ corresponding to each neuron in the ANN is associated with the vector $(\load_1,\ldots,\load_N)$ in the RNN.
In the paper, the authors proposed to use some ``cut--points'' $(\alpha_1,\ldots,\alpha_N)$ such that
$$y_i=0  \Longleftrightarrow \load_i<1-\alpha_i  \qquad \text{ and } \qquad y_i=1  \Longleftrightarrow \load_i>1-\alpha_i.$$
The values of the flow of input signals $\lpi$ and the cut-points $\alpha_i$ must be chosen in order to obtain the expected effect at the output values.
It can be observed that the task of tuning these parameters can be a non trivial problem.

\subsection{Model variations}

There are several extensions of the canonical RNN model. We present a brief summary of a few of them.
\begin{itemize}
\item In~\citep{Gelenbe1993b}, negative arriving customers (signals) make that the
potential of the neurons goes to zero.
\item In~\citep{Gelenbe1999b,Gelenbe2002a,Gelenbe2002c,Fourneau2000250} multiple classes
of neurons are allowed, leading to an increased flexibility in the design
of the model.
\item In~\citep{Gelenbe1993,Gelenbe2000308,Chao1994} negative signal arrivals
can trigger customers' movements in other parts of the network.
\item In~\citep{Henderson1993,Henderson1994,Fourneau95a,Henderson1994b} the rates
of the neurons can be state-dependent, and batch movements are also
allowed.
\end{itemize}

\newcommand{\m}{\mbox{\boldmath{$\mu$}}}
\newcommand{\Pix}{\Prob (C_i|\x)}
\newcommand{\PCi}{\Prob (C_i)}
\newcommand{\pdfxi}{p(\x|C_i)}
\newcommand{\Pjx}{\Prob (C_j|\x)}
\newcommand{\PCj}{\Prob (C_j)}
\newcommand{\pdfxj}{p(\x|C_j)}
\newcommand{\pixbe}{\pi(\x,\bm{\beta})}
\newcommand{\pixbek}{\pi(\xk,\bm{\beta})}

\section{Applications in the supervised learning area}
\label{OverviewRNNsAplications}
Since its apparition in 1989, the RNN model has been applied in a large variety of supervised learning problems.
This article is not a survey about the RNN applications. We focus on the numerical optimization algorithms applied for solving the supervised tasks, as well as in the usage of these algorithms. Both points were already discussed in the previous sections.
However, in the following we briefly describe several applications of the model.
The main references used in this overview are~\citep{Thimotheou10,Kocak11,Kocak00,Sakellari:2010,Do11}.

\subsection{Application for multimedia quality assessment}
Neural Networks have been applied for developing mechanisms of controlling the quality of multimedia applications.
Two main classes of methods can be considered for assessing the perceived quality of multimedia streams:
\begin{itemize}
\item Objective methods:  they are usually based on a comparison between the original and distorted sequences. The main example is the Peak Signal to Noise Ratio (PSNR).
\item Subjective methods: the most commonly used measure is the Mean Opinion Score (MOS), where a group of people evaluate media samples according to a predefined quality scale, under carefully controlled experimental conditions.
\end{itemize}
Thus, the quality is measured as a distance function between the original and the distorted sequences in the case of objective methods, and it is some statistical function based on human evaluations in the case of subjective methods.
These approaches have the following drawbacks.
Objective methods require the original streams which often are not available in practical applications. In particular, this restriction makes that these techniques can not be used for controlling applications, which need real-time reactions.
Subjective methods are expensive and, by definition, they can not be used in real-time problems.
Another remark is that these methods do not take into account the effects of the packet network's parameters in the perceived quality the streams.

A new approach was presented in~\citep{Rubi02:realTimePSQA}, based on subjective techniques and QoS metrics, in order to estimate in real time the quality of the media. We can describe the main idea as follows.
A set of original signals of video or audio (depending of our problem) is considered as the data set.
The training data is generated using a codec and network simulation distortion considering loss rate, delay and other network parameters which affect quality.
Several models to characterize the loss processes in the Internet are used to simulate distorted data, such as independent losses~\citep{Bolot93} or fixed--size loss bursts~\citep{Hands99}, the Gilbert Model~\citep{Gilbert60} and $k^\text{th}$ order Markov chains~\citep{Yajnik99}.
Next, a group of subjects evaluates the training set using a MOS protocol, thus a collection of images or sound distorted is subjectively evaluated.

The RNN has been used for learning the relationship between the network parameters (loss rate, delay, jitter,\dots) and quality as evaluated by panels of human users.
This approach of estimating the multimedia quality the has been studied in the following articles~\citep{Rubi05:PSQA}\citep{Rubi06:PSQA}\citep{Mohamed2004}\citep{Mohamed2004141}, among many others.
In~\citep{Rubi06:PSQA} the model was compared with Naive Bayesian Classifiers and with classic Neural Networks, for this specific supervised learning application the RNN model shows a better performance.

In~\citep{Rubi02:realTimePSQA}, other learning techniques were explored, including those
included in commercial packages. Fig.~\ref{compRubi02} shows one of the situations
where RNN behaved better than a competitor, on the same data and keeping the sizes of
the vectors of weights identical. We can see the classical \textit{overtraining} phenomenon
appearing in the other technique.

\begin{figure}
\centering
\fbox{\includegraphics[width=0.9\textwidth]{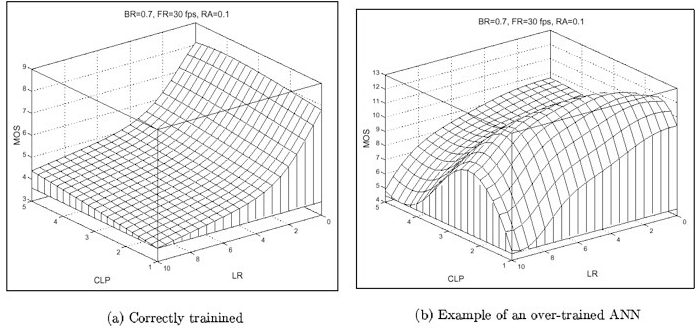}}
\caption{Comparison between the RNN model in (a) and a classical commercial
ANN tool in (b); the picture shows the learned function plotted when varying two of
its input variables. Physical considerations say that the function was corrected
learned in (a), and that (b) is unrealistic.}
\label{compRubi02}
\end{figure}

\subsection{Solving temporal supervised learning}
Recently a computational model that uses the dynamics of the RNN has been analyzed for studying sequential dataset~\citep{Baster12ESQN}\citep{BasterACM13}.
The model name is \textit{Echo State Queueing Network (ESQN)} because it is an hybrid produced from two different types of Neural Networks: RNNs and Echo State Networks (ESN)~\citep{Jaeger09}.
The ESN are recurrent NNs with sigmoid activation functions. They have been successful used for solving time-series problems.
The RNN variation named ESQN uses the circuits of the network for memorizing the sequential data.
Besides, the recurrent part of the network acts as a projection method (as in ESNs and in Kernel Methods) in order to enhance the linear separability of the input data.
The network has fixed the weights that are involved in the circuits and only the weights that generate the output of the model are updated in the training phase.
The state of the nodes in this variation of the RNN model is vector $\bm{\load}$.
When an input pattern $\va^{(k)}$ is presented to the network, the vector of states evolves according to the dynamics given by the following equations, where $\cal I$ denotes the set of input neurons and $\cal H$ denotes the set of hidden ones (the ``reservoir''):

\begin{equation}
\label{ESQNinp}
\load_i^{(k)}=\ds{\frac{a_i^{(k)}}{r_i}}, \quad i \in {\cal I},
\end{equation}
and for $i \in \cal H$,
\begin{equation}
\label{ESQNreservoirState}
    \load_i^{(k)} = \ds{\frac{\ds{\sum_{j\in\mathcal{I}}w^+_{ij}\frac{a_j^{(k)}}{r_j}  + \sum_{j\in\mathcal{I}\cup\mathcal{H}}w^+_{ij}\load_j^{(k-1)}}}%
          {r_i+ \ds{\sum_{j\in\mathcal{I}}w^-_{ij}\frac{a_j^{(k)}}{r_j} + \sum_{j\in\mathcal{I}\cup\mathcal{H}}w^-_{ij}\load_j^{(k-1)}}}},
\end{equation}
for all $i\in\mathcal{H}$.
The expression~(\ref{ESQNreservoirState}) is a dynamical system that has at the left the state values at time~$k$, and on the right hand side the state values at time $k-1$.
In the same way as in the ESN model, the parameters are computed generated using a linear ridge regression from $\bm{\load}$ to $\vb$.

According to the experimental results presented in~\citep{Baster12ESQN} the model presents a competitive accuracy if one refers to the classic Neural Networks with sigmoid activation functions (as, for instance, the ESN model).
In particular, the model has been applied for predicting the Internet traffic on real dataset from an Internet Service Provider (ISP) working in~$11$ European cities~\citep{Cortez12,timesSeries}.
The data was collected from a private ISP from $7$ June to $29$ July, 2005. It covers traffic information of $11$ European cities.
In addition, the model has been applied for traffic prediction using another benchmark data from the \textit{Traffic data from United Kingdom Education and Research Networking Association (UKERNA)}~\citep{Cortez12,timesSeries}.
The data was collected from 19 November and 27 January, 2005.
It was studied in~\citep{Cortez12}.
In order to analyze the sequential behavior, the problem was studied collecting the data and using three-time scales: day, hour and intervals of five minutes.
Different scales of time capture the strength of trend and seasonality.
Figure~\ref{FiguraISP} shows an example of the Internet traffic prediction using the ESQN model on the European ISP dataset. Table~\ref{TablaESQN} illustrates the accuracies of the ESQN and ESN models for predicting the Internet traffic.
For more details about the implementation of ESQNs to solve time-series problems see~\citep{Baster12ESQN}\citep{BasterACM13}\citep{BasterIbica2014}.

\begin{figure*}
\centering
\subfigure[Internet traffic prediction using the RNN model called ESQN on the European ISP traffic validation data ($5$ minutes scale)]
{
\fbox{\includegraphics[width=0.9\textwidth]{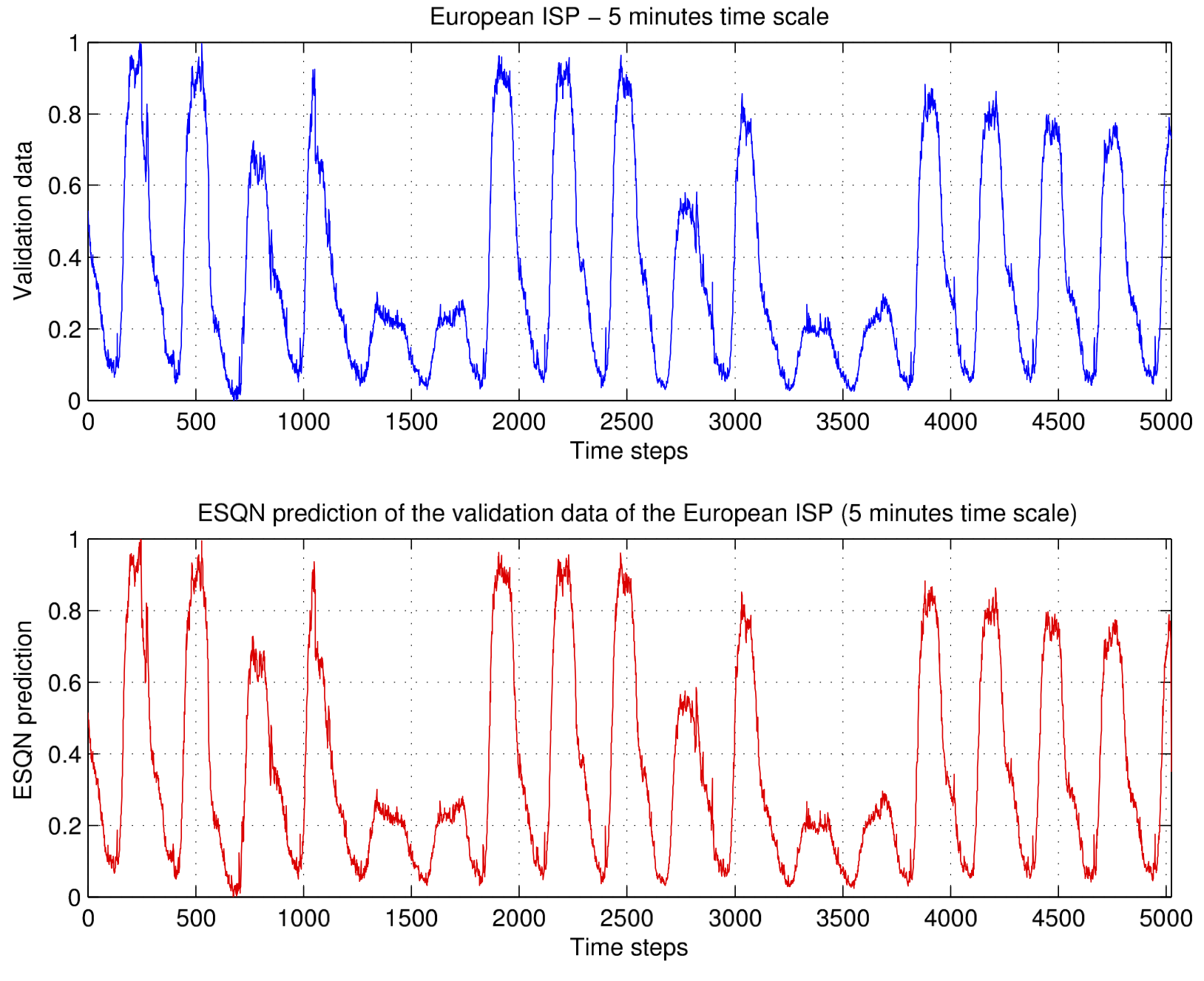}}
\label{EQN/A5MValidation}
}
\subfigure[Internet traffic prediction using the RNN variation called ESQN on the European ISP traffic validation data ($5$ minutes scale). Example of ESQN prediction for $1000$ time steps]
{
\fbox{\includegraphics[width=0.9\textwidth]{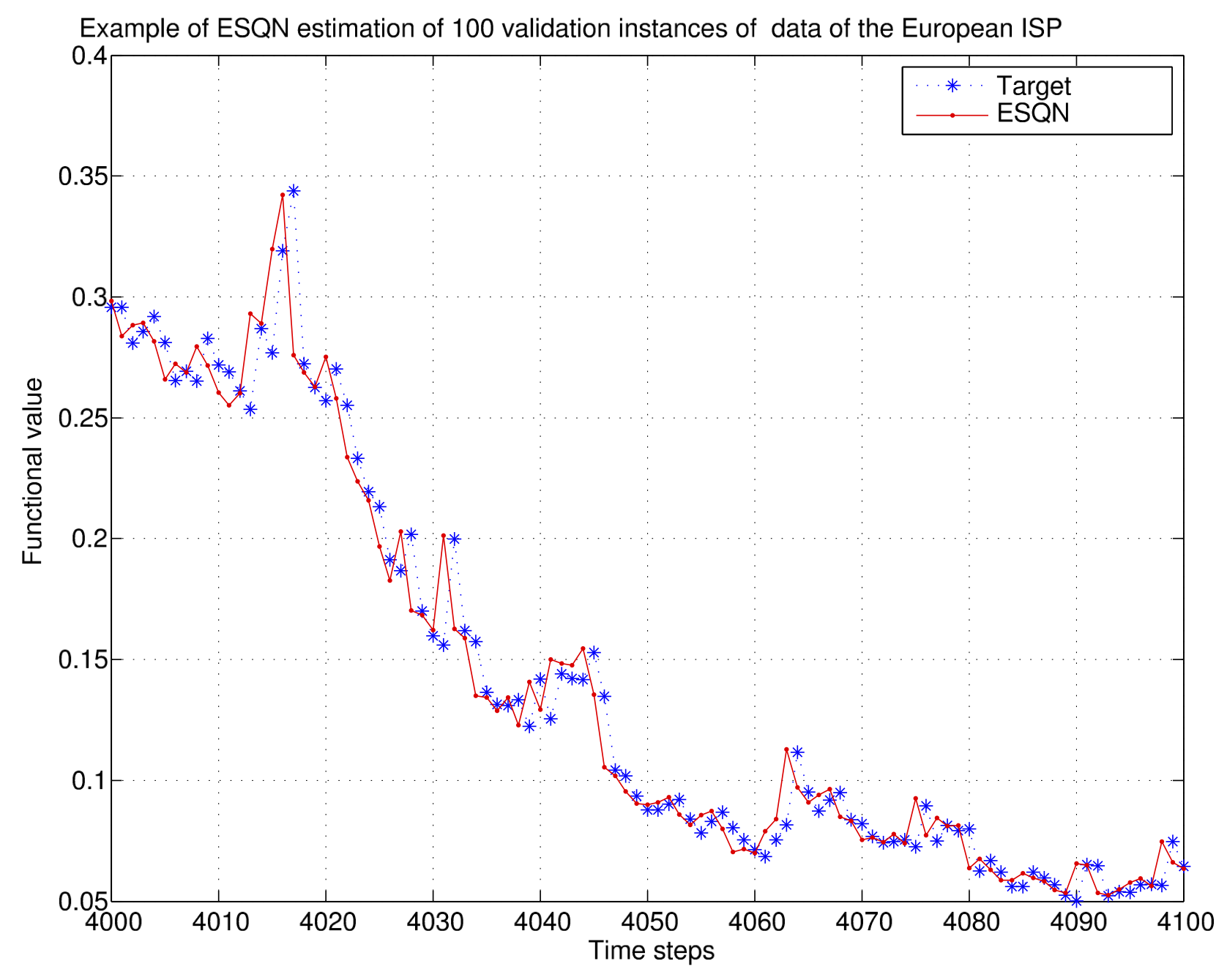}}
\label{ArialZoomA5M_4000_4100}
}
\caption{\label{FiguraISP}Example of Internet traffic prediction using the RNN variation called ESQN on the European ISP traffic validation data ($5$ minutes scale).}
\end{figure*}

\begin{table}[!t]
\centering
\begin{tabular}{cccc}
\hline
\hline
Traffic time series & Model & NMSE & CI\\
\hline
\multirow{2}{*}{ISP}  &  NN variation (ESN) & $0.0062$ & $\pm 9.8885 \times 10^{-7} $\\ 
$\;$ & RNN variation (ESQN) & $0.0100$ & $\pm 1.2436 \times 10^{-4}$\\
\hline
\multirow{2}{*}{UKERNA} & NN variation (ESN) &  $0.3781$ & $\pm0.0066$\\ %
$\;$ & RNN variation (ESQN)  & $0.2030$  & $\pm0.0335$\\ %
\hline
\hline
\end{tabular}
\caption{\label{TablaESQN}Comparison between the ESN (RNN with sigmoid activation function and fixed recurrent structure) and the variation of RNN for time series problems (named ESQN).
The table presents the Normalized MSE reached on~$20$ independent trials, and the corresponding Confidence Interval (CI).
%
}
\label{tablePerformanceChapter4}
\end{table}

\subsection{Approximation of nonlinear functions}
\label{XORproblem}
%
%
%
%
A supervised task consists in generating a parametric mapping between inputs and outputs.
One of the most referenced and studied nonlinear benchmark has been the xor problem.
In 1969, Marvin Minsky and Seymour Papert proved the inability of perceptrons to solve it~\citep{Minsky69}.
Since then, xor is considered as a classical reference to study the ability of a model to solve nonlinear classification problems.
A RNN that tries to match the xor function was studied in~\citep{Gel89:RNNPosNeg,Gel90:stability}.
The same problem solved using learning algorithms was studied in~\citep{Baster09}.
In~\citep{Baster09} a generalization of the xor problem named the parity problem was analyzed.
The RNNs have been used also for approximating real functions.
For instance, in~\citep{Baster09} it was studied for approximating a sinusoidal function.
In~\citep{Martinelli02:Extended} an extension of the canonical RNN has been used for solving real function approximation problems.
%


\subsection{Image processing}
The RNN model has been widely used in image processing problems. In this section we give a brief description about this type of applications.

\subsubsection*{Image compression}
Feedforward NNs have been successful used for compressing images~\citep{Petersen02}.
The network has at least one hidden layer, and the same number of input and output neurons.
The number of hidden neurons is smaller that the number of units in the other layers.
The compression ratio is the rate between the number of input and of hidden neurons.
This specific topology is often referred to as a \textit{bottleneck} network.
We illustrate it in Figure~\ref{Bottleneck}~\citep{Manevitz2007}.
The input data is a collection that represents the original image.
For this purpose, some relevant features of the image is used~\citep{Kocak00}.
In the learning procedure the feedforward network is used as an \textit{auto-associator tool}, which means that the model is trained in order to recreate the input data~\citep{Petersen02}.

The same approach was used with the RNNs on image data compression tasks~\citep{Cramer1996,GelCramerSungur96:traffic,Gel98:ImagComp}.
In~\citep{Kocak00}, RNNs were compared to other more traditional compression tools
from the performance point of view.
The measure of quality considered was the \textit{Peak Signal to Noise Ratio (PSNR)}.
The authors remarked that traditional methods, such as JPG and Wavelet Compression, reached higher performance level (with respect to PSNR) than compression with RNNs.
However, RNN presents the advantage to be faster that the other techniques.
Besides, RNNs for image compression can be adapted for an implementation using parallel computing.
For instance, the image can be fragmented in several parts, and for each part an RNN can be used for compressed/decompressed the image fragments.
This procedure can be implemented in parallel. Thus, the computational time of this compression/decompression technique considerably decreases with respect to the other more classical techniques.

\begin{figure}[h]
\begin{center}
\fbox{\includegraphics[width=0.7\textwidth]{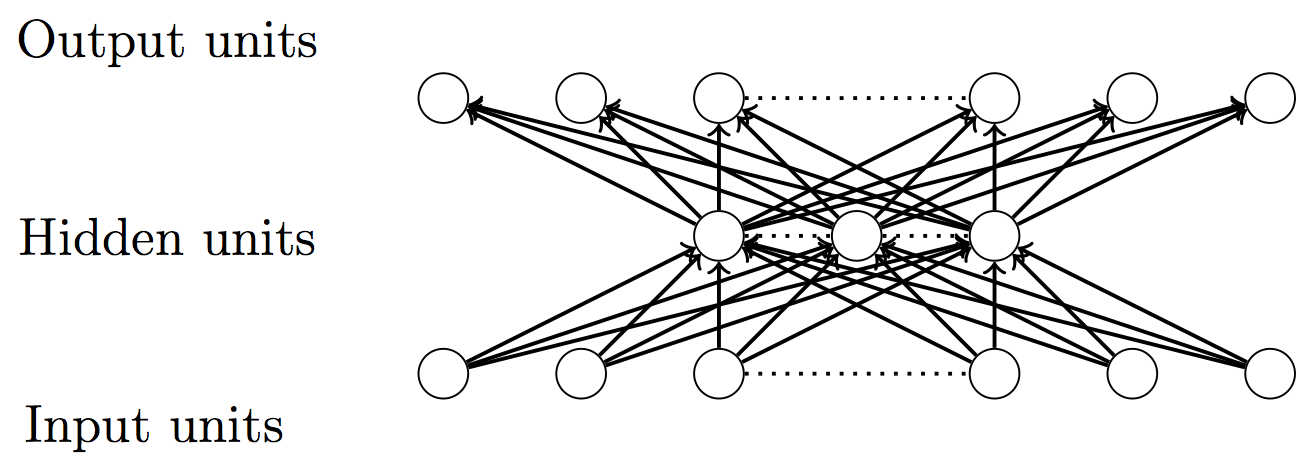}}
\caption[Bottleneck network used for Image Compression]{\label{Bottleneck}Bottleneck network used for compressing images. The network has the same number of input and of output neurons.
The compression ratio is given by ratio between the number of input and of hidden neurons.}
\end{center}
\end{figure}

\subsubsection*{Image enlargement}
The \textit{Image Enlargement} is a technique for resizing a digital image in order to increase its resolution.
%
A procedure to solve this problem using RNNs was developed in~\citep{Baki1998,Kocak00}.
The model was applied for enlarging two well know images called \textit{Lena} and \textit{Peppers}.
The procedure requires a training data composed by pairs of images.
Each pair is composed of the original (the smaller one) and the target (the larger one).
The authors propose to use a feedforward RNN and the gradient descent training algorithm.
The training function is defined using the \textit{zero order interpolation}, a technique for signal reconstruction.
For details about the technicalities of this procedure see~\citep{Petersen02,Kocak00}.
%

\subsubsection*{Image fusion}
In the fusion of images the goal is to obtain a new image with high-resolution from a set of images with low-resolution.
The problem using RNNs was examined in~\citep{Kocak00}.
The training inputs are composed of several images, for instance produced by sensors.
As usual in supervised learning, each input pattern has associated with a target. In this task, the target is an image with better resolution that the one, present at the input.
The network is used for learning the mapping between the set of sensor images and the target image. The authors in~\citep{Kocak00} use Gradient Descent with RNN for solving this problem.
%

\subsubsection*{Medical image computing}

RNNs have been applied in image segmentation for \textit{Magnetic Resonance Imaging (MRI)} of the brain~\citep{GelFeng96:image,Gelenbe1996b}.
The application on MRI is based on the following procedure.
Given a reference image that contains a finite set of regions, the goal is to use RNNs for classifying those regions.
The approach consists of training several RNNs, each of them using a specific region of the reference image.
%
%
Then, after this training phase, an image is decomposed into many blocks of small sizes, and each block is classified using the corresponding RNN.
Thus, we can assign a label to each small block (eventually individual pixels can be considered) of the image.
The model has been applied to the segmentation of ultrasound images~\citep{Lu05}.

\subsubsection*{Textural features for image classification}
\textit{Textural features classification} is an area of image processing and computer vision where the goal is to categorize or to classify pictorial information.
This is done using a set of meaningful features of the image.
%
%
Texture is one of the main characteristics for identifying or classifying individual pixels or pixel blocks in an image.
Several works analyze what kind of textural features are useful for classifying  images~\citep{Haralick73,Rosenfeld70}.
Once the features are defined, the blocks of images can be categorized using some pattern recognition methods.
In~\citep{Teke06:TextureClassif} the authors analyze the performance of using RNNs to classify  pixels into texture classes.
According to this article, the classification performance of RNNs is comparable with that of the other methods presented in the pattern recognition literature.

An extension of RNNs called \textit{Multiple Classes Random Networks}~\citep{Gelenbe1999b} was applied to the Color Pattern Recognition problem in~\citep{Aguilar01}.
The model was also applied to another classification problem called \textit{Laser Intensity Vehicle Classification System (LIVCS)}~\citep{Hussain:2005}.
Basically, laser intensity images are collected in an information system. This kind of images produce information that is less sensitive to environmental conditions that information produced by sensors and video systems. Once the information is collected, specific features are used for training the RNN model.
Another area of application was image reconstruction, see for instance~\citep{Gelenbe2004216}.

\subsubsection*{Other areas of image processing}
In a similar way as for image compression, the RNN model has also been applied to video compression~\citep{GelCramerSungur96:traffic}.
%
%
The authors develop the technique on video sequences over Broadband ISDN networks.
According to~\citep{Cramer1996,Gel98:ImagComp,GelCramer00:video,GelCramerSungur96:traffic}, RNNs exhibit a good performance in these tasks.

\textit{Image synthesis} consists of creating digital images from some texture and form of image description.
The approach using RNNs considers an architecture where each neuron is associated with each pixel of the image.
%
%
To train the networks the method uses a training set composed of images where the aim is to learn certain  textural features such as granularity, inclination, contrast, homogeneity and others.
Once the network is trained, it can be used to generate images with similar textural characteristics as the images used in the learning process.
Models to create image synthesis in gray level using RNNs have been examined in~\citep{GelAtal92:TextureGeneration,GelHusAbdel00:TextureModel,Teke06:TextureClassif}.
There is also some research for color images~\citep{AtalayGel92,Gelenbe2002a}.
\subsection{Other applications of the RNN model}
In addition to supervised learning, the model has been used in the field of combinatorial optimization problems.
Some of the main applications are the following ones (the list is not exhaustive):
\begin{itemize}
\item The \textit{Traveling Salesman Problem (TSP)} is a classical case within hard combinatorial optimization problems.
The goal is to design a set of minimum-cost vehicle routes delivering goods to a set of customers where the vehicles start and finish at a central point.
This problem was studied using RNNs in~\citep{Gelenbe1993d}.
\item Another combinatorial optimization problem is the \textit{Minimum Vertex Covering} (MVC) problem which was studied in~\citep{Aguilar98,Gelenbe1992}.
\item The \textit{satisfiability} (SAT) problem consists of determining if the variables of a given Boolean formula can be assigned in such a way that the formula evaluates to TRUE.
The SAT problem has a great importance in computer science and it was the first known NP--complete problem. It was studied using RNNs in~\citep{Gelenbe1992aa}.
%
\item The well known \textit{Minimum Steiner Tree} (MST) problem in graphs is a sort of benchmark in combinatorial optimization. A RNN-based approach has been developed to find a solution in~\citep{GelenbeGS97,Ghanwani1998}.
\item RNNs have also been used to solve two problems close to MSTs: the \textit{dynamic multicast} problem~\citep{GelenbeGS97,Aiello05} and the \textit{access network design} problem~\citep{Cancela04}.
\item The \textit{Independent Set problem} was studied with a variation of the RNN model in~\citep{Pekergin92}.

\item The \textit{Optimal Resource Allocation} on a distributed system was analyzed using RNNs in~\citep{Zhong05}.
\item  Associative memory: works about associative networks for storage and retrieval symbolic information that use the RNN model can be found in~\citep{Stafylopatis92,Gelenbe91,Gelenbe1991d,Likas97}. In addition, an analogy between the Hopfield Network and the RNN was studied in~\citep{Likas91aninvestigation}.
\item Applications in networking using a routing protocol called the Cognitive Packet Networks are found in~\citep{Liu2007,Gelenbe20071299,Sakellari:2010}.
\end{itemize}

%
%
%
%
%
%
\section{Conclusions and perspectives}
\label{conclusionSection}
Since the early 1990s, the \textit{Random Neural Network (RNN)} model has gained importance in the Neural Networks (NNs) area as well as in the field of the Queueing Theory.
The model can be seen as an extension of open Jackson's networks (it can actually be seen in several different ways).
It has characteristics that are present in biological NNs, such that the action potential, firing spikes among neurons, inhibitory and exhibitory spikes, random delays between spikes, and so on.

This article intends to be a practical guide for applying RNNs in supervised learning tasks.
Several learning algorithms that have been used for training RNNs are presented in some detail.
In 1993, a learning algorithm of the \textit{Gradient Descent (GD)} type was adapted for the RNN case.
At the beginning of the $2000$s, \textit{Quasi-Newton (QN)} methods were applied for training these models.
More recently, other second-order algorithms, namely the \textit{Levenberg-Marquardt (LM)} method and some variations were also developed for training RNNs.
The LM procedure is a more robust technique than Gauss-Newton algorithms and it is very popular in NNs.
In general, the available experimental results have shown that QN methods are much faster than GD algorithms.
However, QN methods (for instance LM) can present robustness problems.
First order optimization models are usually more robust but slower.
In the case of a recurrent topology, a first order method can present also problems of stability. Those drawbacks are identified in the NN literature as the ``exploding'' and ``vanishing'' phenomena.
To the best of our knowledge, those phenomena have not been analyzed yet in the RNN case.

RNNs have been successfully used as learning tools in many applications of different types.
In this overview, we only scratched the surface of some of those applications.
We described some solutions in the area of pattern recognition, image processing, compression, as well as in some combinatorial optimization problems.

Due to the good results previously mentioned, we are encourage researchers to continue the ongoing effort around, in particular, the following lines:
\bit
\item All the learning algorithms presented in the RNN literature use the same type of cost function: a sum of squared ``individual'' errors.
It can be interesting to develop algorithms that use other measures of errors, as for instance the Kullback-Leiber one. It is known that this metric is specially useful for evaluate the learning performance on classification tools~\citep{HastieTibshirami}.
\item A point that deserves specific attention is the study of the
difficulties of training a RNN with a recurrent topology, in particular with
GD-type algorithms. See for instance~\citep{Bengio94,Pascanu13} in the case of
classical NNs. As far as we know, these studies have not been extended yet to
the case of RNNs.
\item There are several optimization algorithms developed in the NN area that are not still studied for the RNN case. For instance, we can mention the Hessian-free optimization and back-propagation curvature~\citep{Martens12ConfICML}.
\item In~\citep{BasterrechCUDA2016,BasterrechCUDA2014} RNNs have been implemented in parallel multiple GPUs. In this implementation, the authors used only the gradient descent method. Therefore, similar works using second-order methods in parallel implementations should be explored as-well.
\eit

\vskip 0.2in
\bibliography{refRnn}

\end{document}